\documentclass{article}

\usepackage[preprint]{neurips_2025}
\usepackage{neurips_2025}
\usepackage{graphicx}
\usepackage{hyperref}
\usepackage{defs}
\usepackage{fullpage}
\usepackage{algorithm,algorithmicx}
\usepackage{algpseudocode}
\usepackage{natbib}
\usepackage{cases}
\usepackage{ulem}
\usepackage{xcolor}
\usepackage{enumitem}
\usepackage{tcolorbox}
\usepackage{adjustbox}
\usepackage{subcaption}
\usepackage{ragged2e}
\usepackage{enumitem}
\usepackage{pgfplots}
\pgfplotsset{compat=1.17}
\tcbuselibrary{breakable}

\definecolor{nmgray}{RGB}{229,229,229}

\usepackage[utf8]{inputenc} 
\usepackage[T1]{fontenc}    
\usepackage{hyperref}       
\usepackage{url}            
\usepackage{booktabs}       
\usepackage{amsfonts}       
\usepackage{nicefrac}       
\usepackage{microtype}      
\usepackage{xcolor}         
\usepackage{ulem}

\title{Improving the Data-efficiency of Reinforcement Learning by Warm-starting with LLM}

\author{%
  Thang Duong \\
  University of Arizona \\
  \texttt{thangduong@arizona.edu} \\
  \And
  Minglai Yang \\
  University of Arizona \\
  \texttt{mingly@arizona.edu} \\
  \And
  Chicheng Zhang \\
  University of Arizona\\
  \texttt{chichengz@cs.arizona.edu} \\
}

\newcommand{\promptbox}[2]{%
  \begin{tcolorbox}[
    breakable,
    title=#1,
    colback=gray!5,
    colframe=nmgray!75!black,
    before skip=1pt,
    after skip=1pt,
    fontupper=\linespread{0.8}\selectfont]
    #2
  \end{tcolorbox}}
\begin{document}

\maketitle

\begin{abstract}
  We investigate the usage of Large Language Model (LLM) in collecting high-quality data to warm-start Reinforcement Learning (RL) algorithms for learning in some classical Markov Decision Process (MDP) environments. In this work, we focus on using LLM to generate an off-policy dataset that sufficiently covers 
  state-actions visited by optimal policies, then later using
  an RL algorithm to explore the environment and improve the policy suggested by the LLM. Our algorithm, LORO, can both converge to an optimal policy and have a high sample efficiency thanks to the LLM's good starting policy. 
  On multiple OpenAI Gym environments, such as CartPole and Pendulum, we empirically demonstrate that LORO outperforms baseline algorithms such as pure LLM-based policies, pure RL, and a naive combination of the two, achieving up to $4 \times$ the cumulative rewards of the pure RL baseline.
%
\end{abstract}

\section{Introduction}

The standard protocol in 
online RL has many interesting applications, from playing games \cite{silver2017mastering} to robotic control \cite{kober2013reinforcement}. While having impressive empirical performance and enjoying the theoretical guarantee on returning the optimal policy under some assumptions \cite{ramaswamy2021deepqlearningtheoreticalinsights}, \cite{agarwal2019reinforcement}, \cite{bertsekas2007neuro}, a key problem of this approach is the notorious sample inefficiency, which limits its application in practice. Thus, most impressive successes in online RL have been restricted to settings where many samples can be obtained by interacting with the environment (such as games or environments with high-quality simulations).

To overcome this, \cite{lange2012batch}, \cite{ernst2005tree}, \cite{riedmiller2005neural}, and \cite{levine2020offline} proposed the Offline RL setting, where the algorithm does not directly interact with the environment as in online RL, but is trained on a large dataset of experience collected from some other sources (e.g., by expert demonstration). While the sample complexity problem is mitigated due to the large training dataset, these Offline RL methods suffer from the \textit{distribution shift} problem, where the state distribution from the offline data differs significantly from the one induced by online interactions \cite{wang2021instabilities}.

A popular approach to address the distribution shift problem is by aggregating both the offline and online data \cite{xie2021policy}, \cite{song2022hybrid}, \cite{zhang2023policy}
This offline-to-online approach greatly reduces the sample complexity of the problem by reducing unnecessary exploration with the offline dataset while also mitigating the distribution shift problem through online interactions. Under some assumptions, \cite{song2022hybrid} provides a cumulative regret 
and sample complexity 
guarantee for the offline-to-online setting. They show that, if the offline data distribution covers some high-quality policies' trajectories, their offline-to-online algorithm is both sample-efficient and competitive with the high-quality policies covered by the offline data. 
Even then, our goal is to further improve the sample efficiency, perhaps by leveraging extra information from the problem description and common sense (e.g., avoid obstacles, find the key to open the door, etc.)

Recently, LLM  has shown a remarkable ability for reasoning, even in hard sequential decision problems such as robot manipulation \cite{ahn2022can}, \cite{huang2022inner}, \cite{liang2023code}.
Still, \cite{carta2023grounding} claims that, while useful, LLM's application for sequential decision tasks is limited by the absence of environment grounding. They demonstrate this in an interactive textual environment and show that online interaction with the environment can help "ground" the LLM and boost the overall performance.
Thus, we raise the question:

\begin{center}
    \textit{Can we design an algorithm that can both converge to the optimal policy and has high sample efficiency by leveraging the reasoning capability of LLMs?}
\end{center} 

In this paper, we answer this question positively. Under Assumption \ref{assum:coverage}, where the policy suggested by the LLM has sufficient coverage of an optimal policy,
 our algorithm, LLM Off-policy pretrain, RL On-policy (LORO), enjoys both small Cumulative suboptimality and Sample complexity as suggested by \cite{song2022hybrid}. To the best of our knowledge, we are the first to suggest warm-starting RL with LLM's collected data and connect LLM with offline-to-online RL, drawing the similarity between the distribution shift problem in Offline RL versus the useful-but-suboptimal policy extracted from LLM, and suggest that the offline-to-online RL approaches can be applied here. Empirically, we demonstrate the effectiveness of our algorithm in multiple OpenAI Gym environments \cite{towers2024gymnasium} compared to other baselines such as pure LLM-based policies, pure RL, and a naive combination of the two.

\section{Related work}

\textbf{Offline-to-online RL:} \cite{nair2020awac} showed that a naive combination of offline pre-training and online fine-tuning does not usually help and often worsen the performance, a large part due to excess conservatism. Many more sophisticated approaches have been studied empirically: \cite{nair2018overcoming}, \cite{hester2018deep}, \cite{uehara2021representation}, \cite{ball2023efficientonlinereinforcementlearning}. However, none of these works studied the utility of LLMs in warm-starting online RL. In contrast, our goal is to improve the sample efficiency further by combining this approach with LLM, especially if the LLM can understand the problem description and use common sense to take good actions. 

\textbf{Warm-starting RL}: 
\cite{schmitt2018kickstarting} propose to kick-start
Deep RL with a teacher policy by adding an extra objective to encourage the learner to behave similarly to the teacher, with a diminishing weight to allow the student to eventually surpass the teacher. One limitation is that \cite{schmitt2018kickstarting} assumes the teacher policy is high-performing enough 
enough to be distilled, meaning its application is limited when learning a new task from scratch. In contrast, we only require the initial policy to sufficiently cover the state-action pairs often visited by an optimal policy. This is a much milder assumption and is reflected in our Experiment section, where a very weak initial LLM policy can still be useful. 
We are also focusing more on leveraging the LLM capability to zero-shot boost the performance of RL tasks, which is critical in problems with high labeling cost or that have safety as a main concern, which \cite{schmitt2018kickstarting}'s approach may struggle with their requirement of a strong initial teacher policy.


\textbf{Embodied LLM and environment interactions:} Recently, LLM has showed very impressive capability \cite{brown2020language}, including understanding about physics \cite{patel2022mapping} \cite{liu2024physgen}, color \cite{abdou2021can}, and affordances between bodies and object \cite{ahn2022can}. This implicit knowledge could be the reason why LLM can be used to directly manipulate robots \cite{ahn2022can}, \cite{huang2022inner}, \cite{liang2023code}. However, \cite{carta2023grounding} claims that LLMs lack grounding due to 1) the training objective of next word prediction not aligned with other goals, and 2) no interactions with the environment. 
%

Many works seemingly agree with \cite{carta2023grounding} and incorporate environment interactions, thus showing significant improvement. A popular approach is letting the LLM interact directly with the environment and collect the feedback for the subsequent prompt \cite{carta2023grounding}, \cite{yao2022react}, \cite{zhou2023language}, \cite{luketina2019survey}. Another direction is a two level system, where the LLM take high level, abstract actions (such as creating sub-goals \cite{bhat2024grounding} \cite{dalal2024plan} or choosing the skills to use \cite{liang2023code}, \cite{ahn2022can}), and the low level classical system implementing the LLM's ``plan'' in practice. A related work from \cite{hao2024planning} uses LLM to extract and formulate the problem's objectives, constraints, and may include sub-goals creation, for the low-level optimization solver.

\textbf{Theoretical analysis on LLM's exploration in MDPs:} Many works investigate how LLM performs compared to traditional methods, such as UCB, in MDP problems. \cite{arumugam2025efficientexplorationlargelanguage} introduces a more explicit method for exploration using Posterior Sampling. \cite{chen2024efficient} uses LLM to construct multiple policies and combine with a model selection algorithm to solve Contextual Bandit. \cite{nie2024evolve}, \cite{krishnamurthy2024can}
investigate how LLM explores in the Bandit problem and show that the base LLM policies are non-trivial, but sub-optimal. This assessment aligns with our experiment results.

\textbf{Using LLM to provide extra information for RL: } \cite{carta2023grounding} and \cite{tan2024true} use LLM directly to generate the policy and fine-tune it with RL (using Policy Gradient with PPO or an Actor-Critic framework). \cite{lee2023supervised} and \cite{lin2023transformers} propose pretraining an LLM with an offline dataset and show that it can both explore online and act conservatively offline. Unlike them, instead of an end-to-end approach that mixes the RL objective (of maximizing the cumulative reward) with the LLM objective (for next token prediction), we have a separate, smaller RL learner trained exclusively on the classical RL objective that enjoys the typical asymptotic optimality 
Since one of our motivations is computational efficiency, hence, training a large neural network that requires a lot of data would defeat the point of using LLM to help reduce the sample complexity. \cite{pmlr-v162-zheng22c} pre-trains a transformer-based neural network on the offline dataset and develops a way to efficiently fine-tune it with online interaction. This differs from our proposal since we don't have an offline dataset, but the data collected by the LLM's policy can be regarded as a small offline dataset. Another closely related work is \cite{du2023guiding}, where the LLM guides the algorithm's exploration by generating (sub) goals and rewards the RL algorithm when achieving these goals. While both this work and ours leverage LLM to reduce unnecessary exploration for RL, they focus more on sub-goal generation and providing intrinsic reward in sparse feedback problems, while we are focusing on dense reward settings where RL online interactions can refine the warm-started but sub-optimal policy given by the LLM. Finally, \cite{choi2022lmpriors} and \cite{kant2022housekeep} use LLM to provide a prior for the policy to help the learner explore more efficiently, which is similar to our motivation on a high level.

\textbf{Other ways LLM can help solving MDPs:} Besides low-level control and high-level planning, \cite{app14198868} also investigates how LLM can help robot intelligence systems by reward design (to combine with RL) \cite{ma2023eureka}, \cite{xie2023text2reward}, and scene understanding \cite{huang2023visual}, \cite{hong20233d}. Even with these successes, there are still many challenges in deploying LLM to solve sequential decision problems in practice, such as the lack of a guarantee of finding the optimal solution. 



\section{Preliminaries}

Consider a Markov Decision Process $M(\Scal, \Acal, T, H, R, P, d_0)$, where $\Scal$ is the state space, $\Acal$ is the action space, $T$ is the number of episode, $H$ is the horizon of each episode, the reward function is $R(s, a) \in \Delta([0,1])$ and the transition dynamic $P(s,a) \in \Delta(\Scal)$ at $(s, a)$, and $d_0(\Scal)$ is the initial distribution. At each step $h$, the learner chooses from its policy $\pi$ an action $a_h \leftarrow \pi(s_h)$ and receive the reward from the Reward function: $r_h \leftarrow R(s_h, a_h)$,
and transitions to the next state $s_{h+1} \sim P(s_h, a_h)$
. The optimal policy $\pi^*$ is defined as a policy that has a maximum expected cumulative reward: $\pi^* = \argmax_{\pi} \EE\sbr{ 
\sum_{h=1}^H r_h \mid \pi }$.


We also have access to an initial policy $\pi_{\mathrm{LLM}}$
that satisfies Assumption \ref{assum:coverage}. The goal is to maximize the cumulative reward by making use of $\pi_{\mathrm{LLM}}$ to improve the sample efficiency.

\begin{assum}
    Following $\pi_{\mathrm{LLM}}$ can produce trajectories with state-action pairs that sufficiently cover an optimal policy of the MDP.
    \label{assum:coverage}
\end{assum}






Assumption~\ref{assum:coverage} states that an LLM can zero-shot suggest non-trivial base policies, but they are not optimal. 
We see an analogous phenomenon with distribution shift in offline RL that results in a suboptimal policy.
%
%
Thus, we hypothesize that aggregated trajectories collected with LLM, which avoid trivial state-action data (such as unnecessary repetitions, visiting absorbing states, etc), and refine the learned policy later with online interaction, as the offline-to-online protocol, can be useful. Hence, we create Assumption \ref{assum:coverage}, which enables the LLM's collected dataset to satisfy \cite{song2022hybrid}, thus allowing our algorithm to enjoy their Cumulative suboptimality regret and Sample complexity guarantees.

\section{The LLM Off-policy pre-train, RL On-policy (LORO) Algorithm}

Our LORO algorithm is very straightforward. Under Assumption \ref{assum:coverage}, the policy $\pi_{\mathrm{LLM}}$ collects high-quality data from the region that an optimal policy often visits. By only focusing on this and not exploring the low-quality data regions that are avoided by all optimal policies (\textit{e.g.}, hitting the wall, absorbing states, etc.), we can significantly increase the sample efficiency. Thus, we use $\pi_{\mathrm{LLM}}$ to collect a small amount of data, off-policy "pre-train" a policy $\pi$ with it, and then do on-policy online learning to fine-tune $\pi$ to be optimal with a much smaller number of observations.

\begin{figure}[h]
\centering
\includegraphics[scale=0.83]{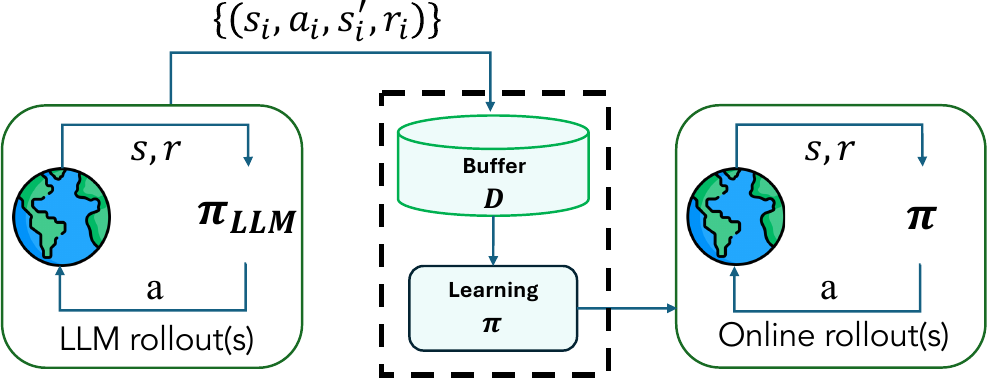}
\caption{The LLM Off-policy pre-train, RL On-policy (LORO) algorithm. Image inspired by \cite{levine2020offline}. }
\label{fig:loro}
\end{figure}

The details of our algorithm, LORO, are shown in Figure \ref{fig:loro} and Algorithm \ref{alg:loro}. 
Initially, we use the LLM policy $\pi_{\mathrm{LLM}}$ to collect data for the first $\tau$ episodes (line \ref{alg:loro-0} - \ref{alg:loro-1}). Then, we pre-train a policy using a classical RL algorithm on the data collected by LLM (line \ref{alg:loro-2}). Finally, we online fine-tune the pre-trained policy (line \ref{alg:loro-3} - \ref{alg:loro-4}).


\begin{algorithm}[t]
    \caption{LLM Off-policy pre-train, RL On-policy (LORO)}
    \begin{algorithmic}[1]
    
    \State \textbf{Input:} $\#$ of episodes $T$, $\#$ of LLM data collection episode $\tau$, episode length $H$, off-policy RL algorithm $\mathrm{Alg}(\cdot)$

    \State \textbf{Initialize:} LLM off-policy dataset: $\Dcal = \cbr{\emptyset}$
    \For{$t = 1, \cdots, \tau$} \label{alg:loro-0} \Comment{LLM data collection} 
        \For{$h = 1, \cdots, H$}
            \State Observe state $s^t_h$, take action $a^t_h \leftarrow \pi_{\mathrm{LLM}}(s^t_h)$, and receive reward $r^t_h$
            \State Aggregate data $\Dcal \leftarrow (s^t_h, a^t_h, r^t_h)$
        \EndFor
    \EndFor \label{alg:loro-1}
    \State Pre-train the policy $\pi_{prev} \leftarrow \mathrm{Alg}(\Dcal)$ \label{alg:loro-2}
    \For{$t = \tau+1, \cdots, T$} \label{alg:loro-3} \Comment{Online learning}
        \For{$h = 1, \cdots, H$}
            \State Get policy $\pi^t_h \leftarrow \mathrm{Alg}(\Dcal, \pi_{prev})$ \Comment{Online updating the policy with the new data}
            \State Observe state $s^t_h$, take action $a^t_h \leftarrow \pi^t_h(s^t_h)$, receive reward $r^t_h$
            \State Aggregate data $\Dcal \leftarrow \Dcal \cup \cbr{(s^t_h, a^t_h, r^t_h)}
            $
            \State Update $\pi_{prev} = \pi^t_h$
        \EndFor
    \EndFor \label{alg:loro-4}
    \end{algorithmic}
    \label{alg:loro}
\end{algorithm}

\section{Experiment}

We empirically evaluate our algorithm\footnote{The code of our experiment can be viewed at \url{https://github.com/duongnhatthang/LlamaGym}} on a host of RL environments: Cart Pole, Pendulum, Frozen Lake, Cliff Walking, Represented Pong, and Mountain Car. 
We defer the environments' descriptions and RL implementation details to Appendix \ref{sec:appendix_env_impl}, and the LLM setup to Appendix \ref{appendix:llm_setup}.

Here, we compare our algorithm with these baselines:
\begin{itemize}
    \item \textbf{On-policy}: a classical on-policy RL method, in which the algorithm collects the data and refines its policy in an online manner.
    \item \textbf{LLMs as Policies (Qwen-7B-Instruct, Qwen-32B-Instruct)}: 
    the base policies from the 7B and 32B of the Qwen 2.5 series with Instruction tuning \cite{yang2024qwen2}.
    For each episode $t$ and step $h$, the LLM has access to the environment and observation 
    descriptions $s^t_h$, and the action $a^t_h$ is taken using Chain-of-Thought \cite{wei2022chain}. The LLM setup details are in Appendix \ref{appendix:llm_setup}. The prompt setup and examples are in Appendix \ref{appendix:prompt}. Note that we only show the average episode reward collected in the first $\tau$ episodes, $r_{\mathrm{avg}} = \frac{1}{H \tau} \sum_{t=1}^\tau \sum_{h=1}^H r^t_h$, in the figures below.
    \item \textbf{Random}: a policy $\pi_{\mathrm{random}}$ that take uniformly random action $a^t_h$. Similarly, we only show the average episode reward collected in the first $\tau$ episodes.
\end{itemize}

In the experiments below, we choose $\tau=10$ and the number of pre-training steps is 1000. The task length $T$ is 150 for CartPole, FrozenLake, 200 for CliffWalking, Pendulum, RepresentedPong, and 300 for MountainCar. LORO is trained using the data collected by Qwen-7B. 
To thoroughly evaluate the effectiveness of LORO and understand the underlying factors influencing its performance, we organize our experiments into several parts: we begin by demonstrating the main results (§\ref{subsec:main}) across diverse RL environments, then conduct a series of ablation studies to assess the impact of pre-training (§\ref{subsec:effect_pretraining}), the quality of LLM-collected data (§\ref{subsec:effect_llmdata}), and the LLM’s reasoning capabilities (§\ref{subsec:effect_llm_capability}).
In addition, we explore the effects of LLM model size, and the number of pre-training steps and varying the amount of LLM data $\tau$, with results provided in Appendix~\ref{appendix:model_size}, \ref{appendix:pre_step}, and \ref{appendix:pre_eps}.



\subsection{Main results}
\label{subsec:main}

\begin{figure}[hbtp!]
\captionsetup[subfigure]{justification=Centering}

\begin{subfigure}[t]{0.48\textwidth}
    \includegraphics[width=\textwidth]{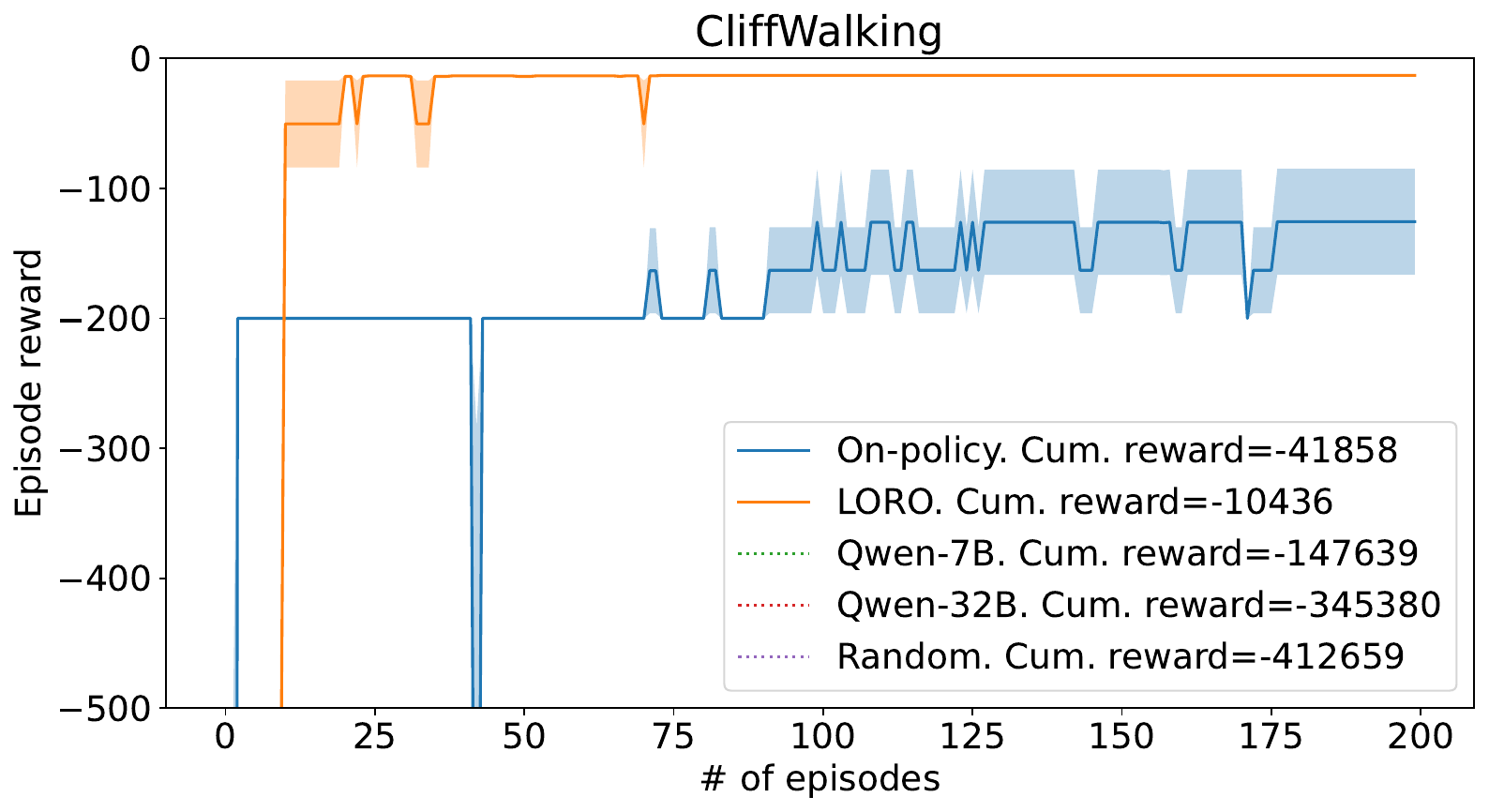}
\end{subfigure}\hspace{\fill} 
\begin{subfigure}[t]{0.48\textwidth}
    \includegraphics[width=\linewidth]{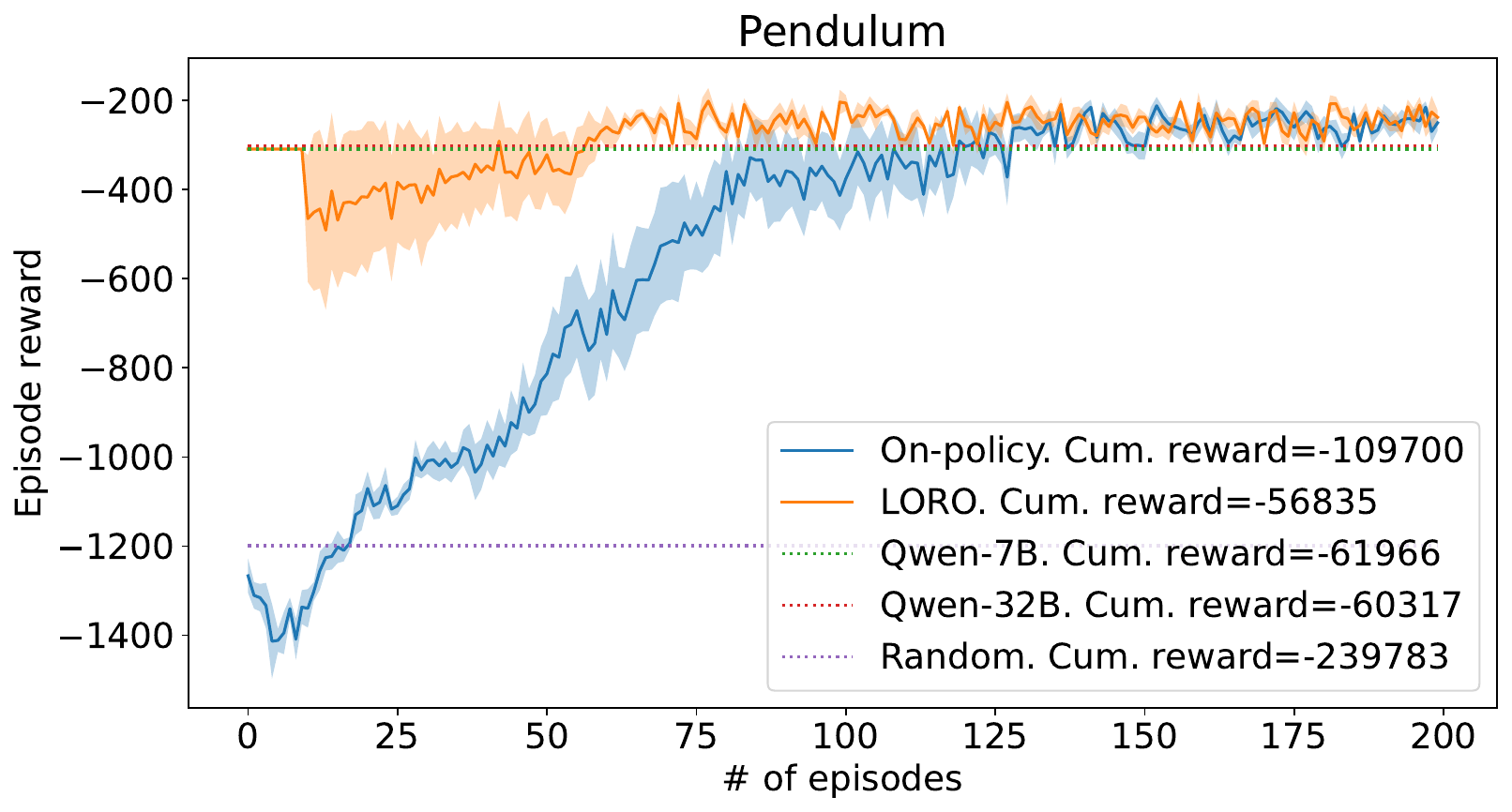}
\end{subfigure}

\begin{subfigure}[t]{0.48\textwidth}
    \includegraphics[width=\textwidth]{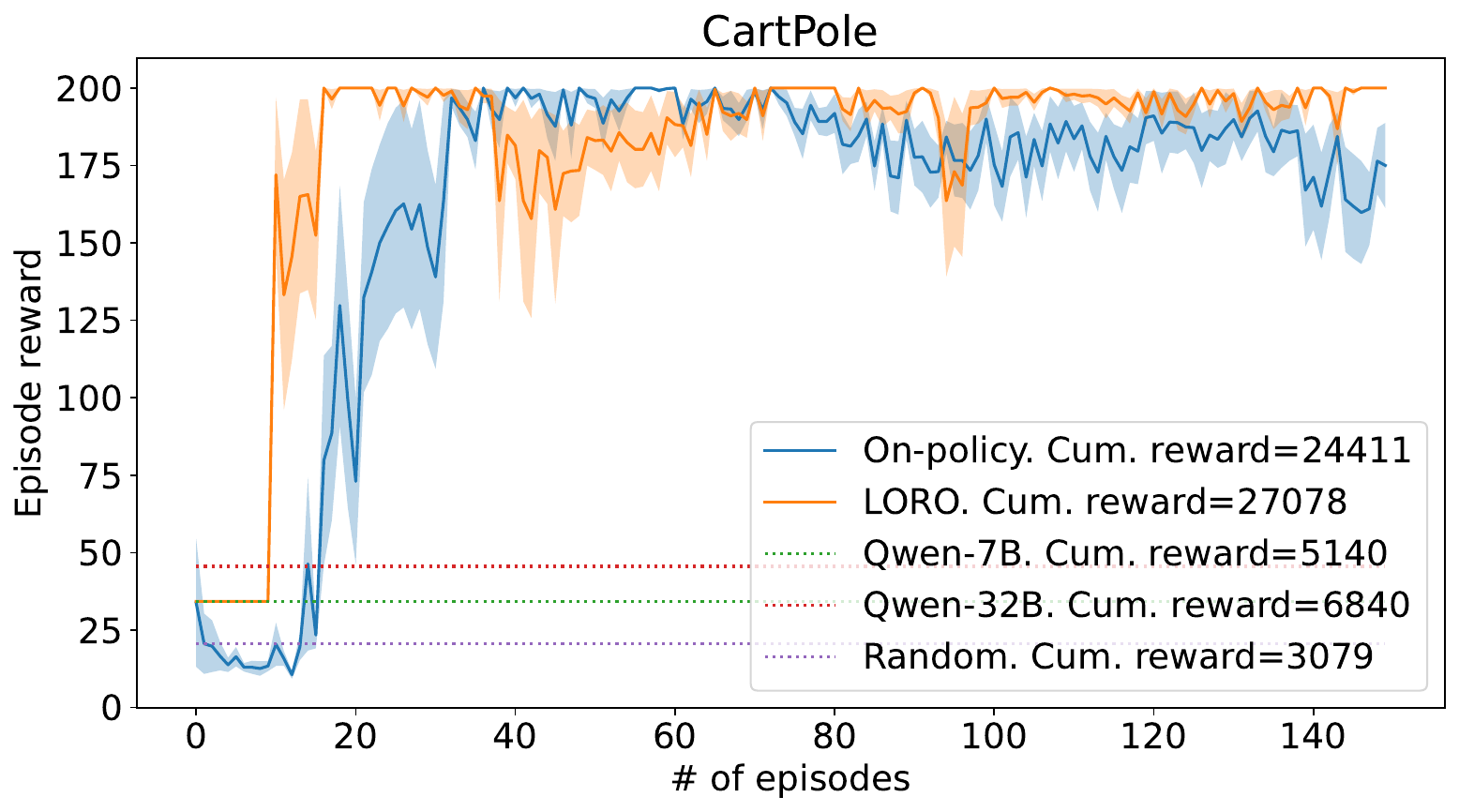}
\end{subfigure}\hspace{\fill} 
\begin{subfigure}[t]{0.48\textwidth}
    \includegraphics[width=\linewidth]{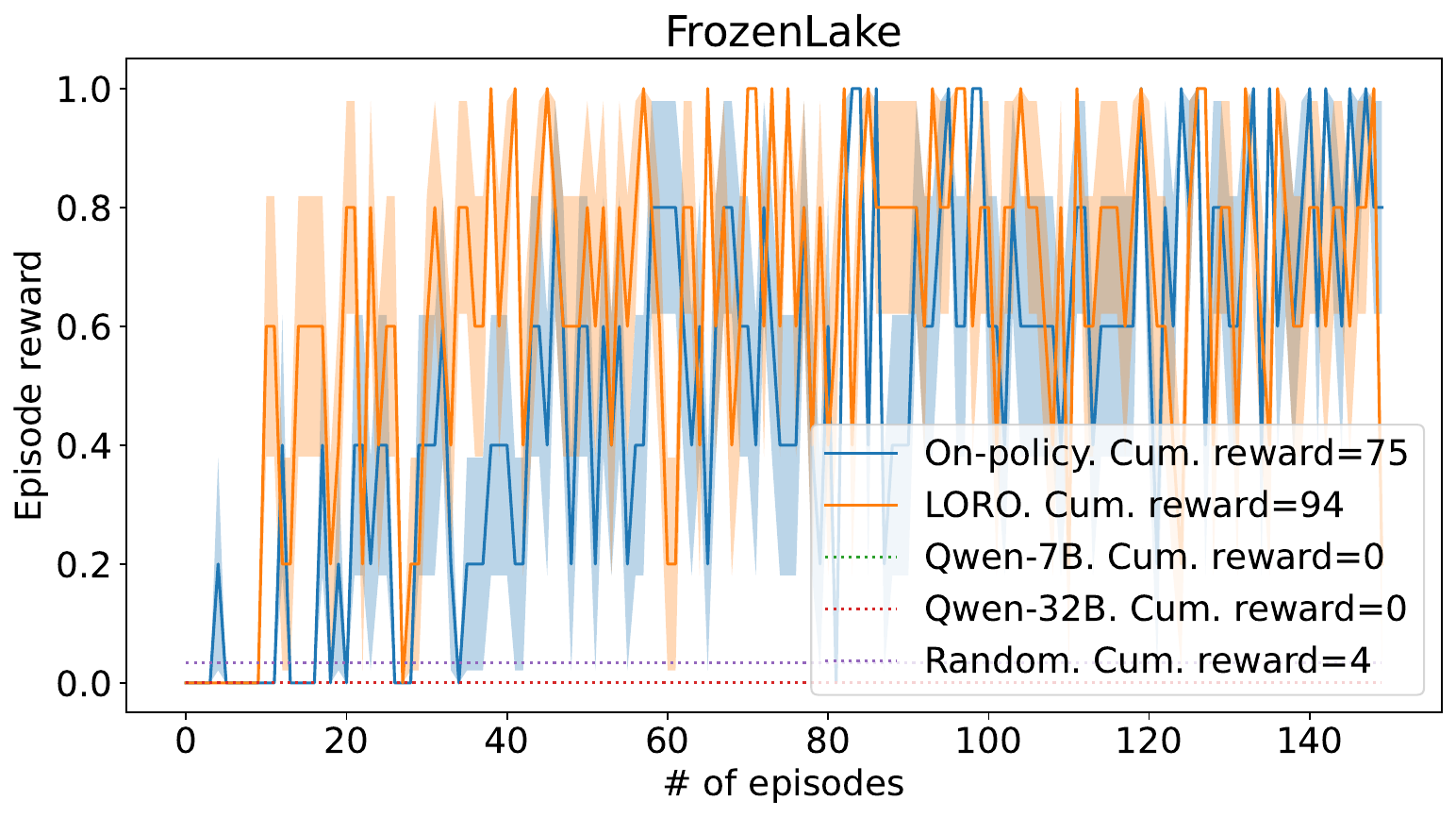}
\end{subfigure}

\begin{subfigure}[t]{0.48\textwidth}
    \includegraphics[width=\linewidth]{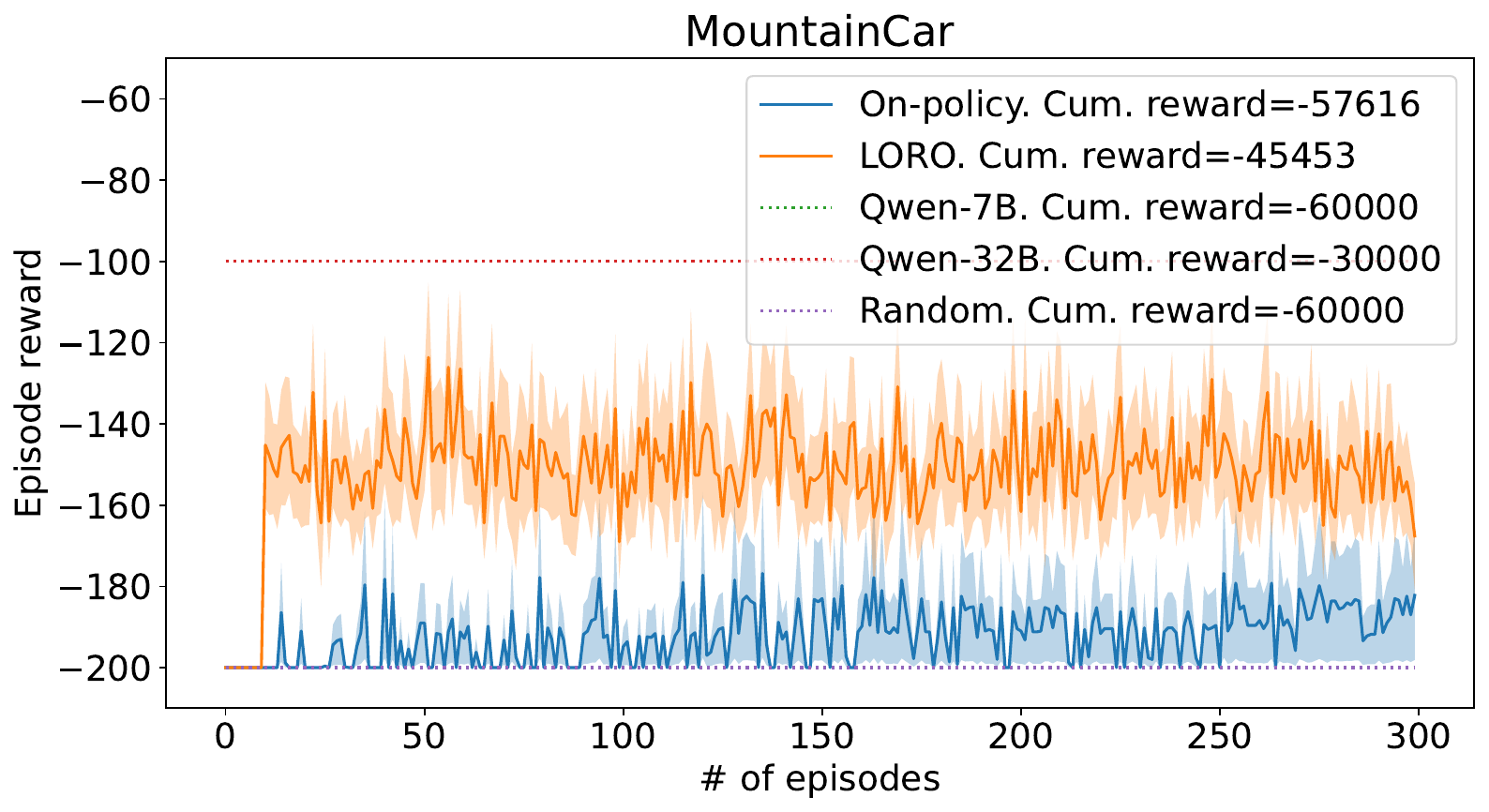}
\end{subfigure}\hspace{\fill} 
\begin{subfigure}[t]{0.48\textwidth}
    \includegraphics[width=\linewidth]{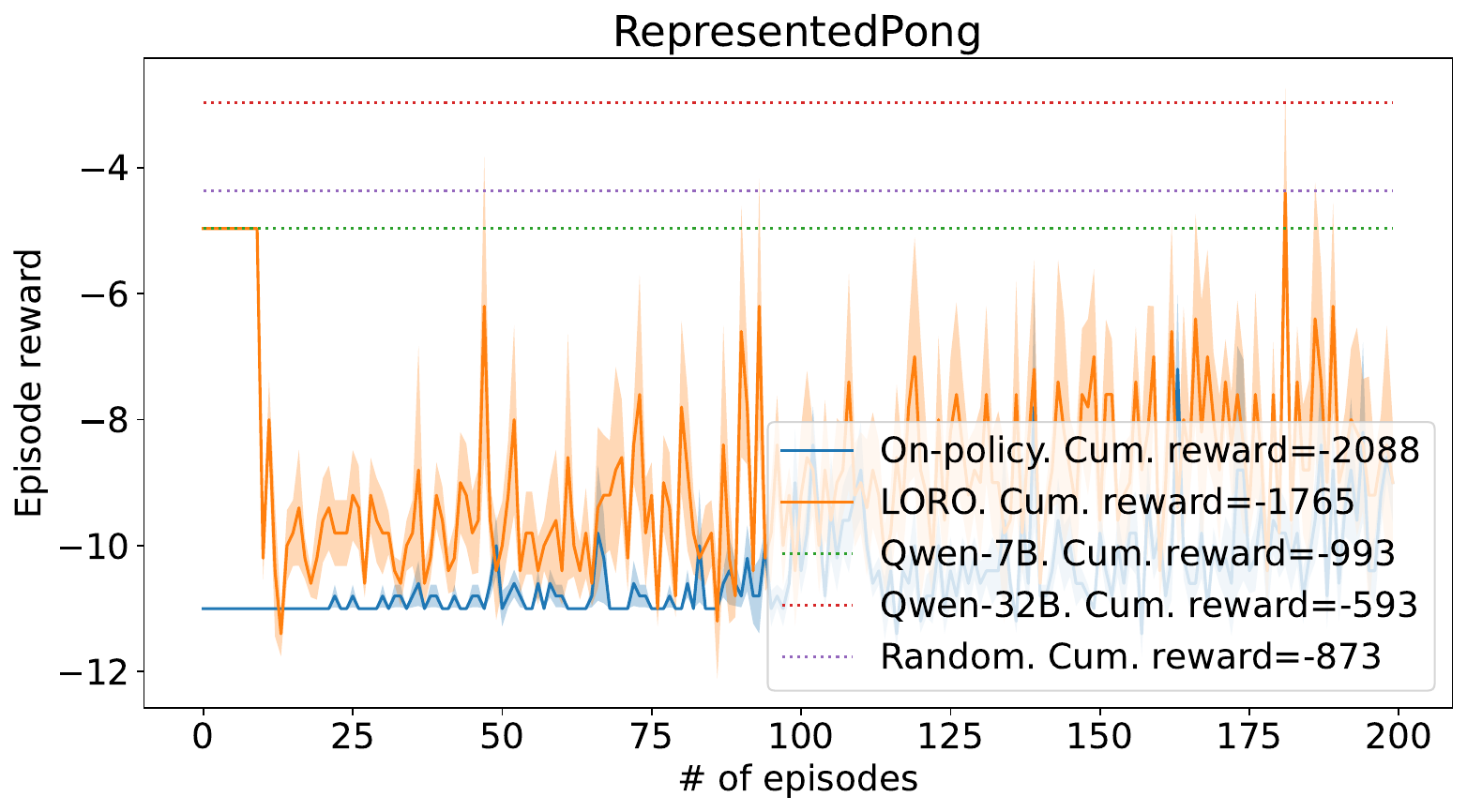}
\end{subfigure}
\caption{Our algorithm, LORO, outperforms the LLM policies (Qwen 7B, Qwen 32B) and the vanilla On-Policy RL baselines. We demonstrate the effectiveness of LORO in six environments, reducing the data required to learn the optimal policy from two to ten times the vanilla On-Policy baseline. LORO and the On-policy baseline learn the optimal policy in the first four environments. Even when not converged to the optimal solution, LORO outperforms other baselines in the last two environments. LLM's base policies can perform very well in the Pendulum experiment, maybe because of the extra information that we provide for this setting, described in Appendix \ref{sec:appendix_env_impl}. All results are shown with standard error over five random seeds. In the CliffWalking experiment, some baselines are not shown in the figure since their episode rewards are too small. Similarly, multiple baselines overlap at -200 on the MountainCar experiment.}
\label{fig:main}
\end{figure}

The main results of our algorithm are shown in Figure \ref{fig:main}. 
In all learning curves, the first $\tau=10$ episodes in LORO showed the average episode reward using the pure LLM-based policies. Afterwards, LORO significantly outperforms the LLM-based policies and the On-Policy RL baselines. Notice that the base LLM policies are often not optimal, but they can still generate high-quality trajectories to improve the sample efficiency of LORO, which justifies Assumption \ref{assum:coverage} and \cite{song2022hybrid}'s claim.

\subsection{Effect of pre-training}
\label{subsec:effect_pretraining}


We would like to verify 
the importance of pre-training (Algorithm \ref{alg:loro} line \ref{alg:loro-2})
is for LORO's performance through an ablation study, since \cite{song2022hybrid}'s algorithm only mixes 
the offline and online
data without pre-training to avoid being too conservative toward the pre-training data, which may hurt the performance.

\begin{figure}[hbtp!]
\captionsetup[subfigure]{justification=Centering}

\begin{subfigure}[t]{0.48\textwidth}
    \includegraphics[width=\textwidth]{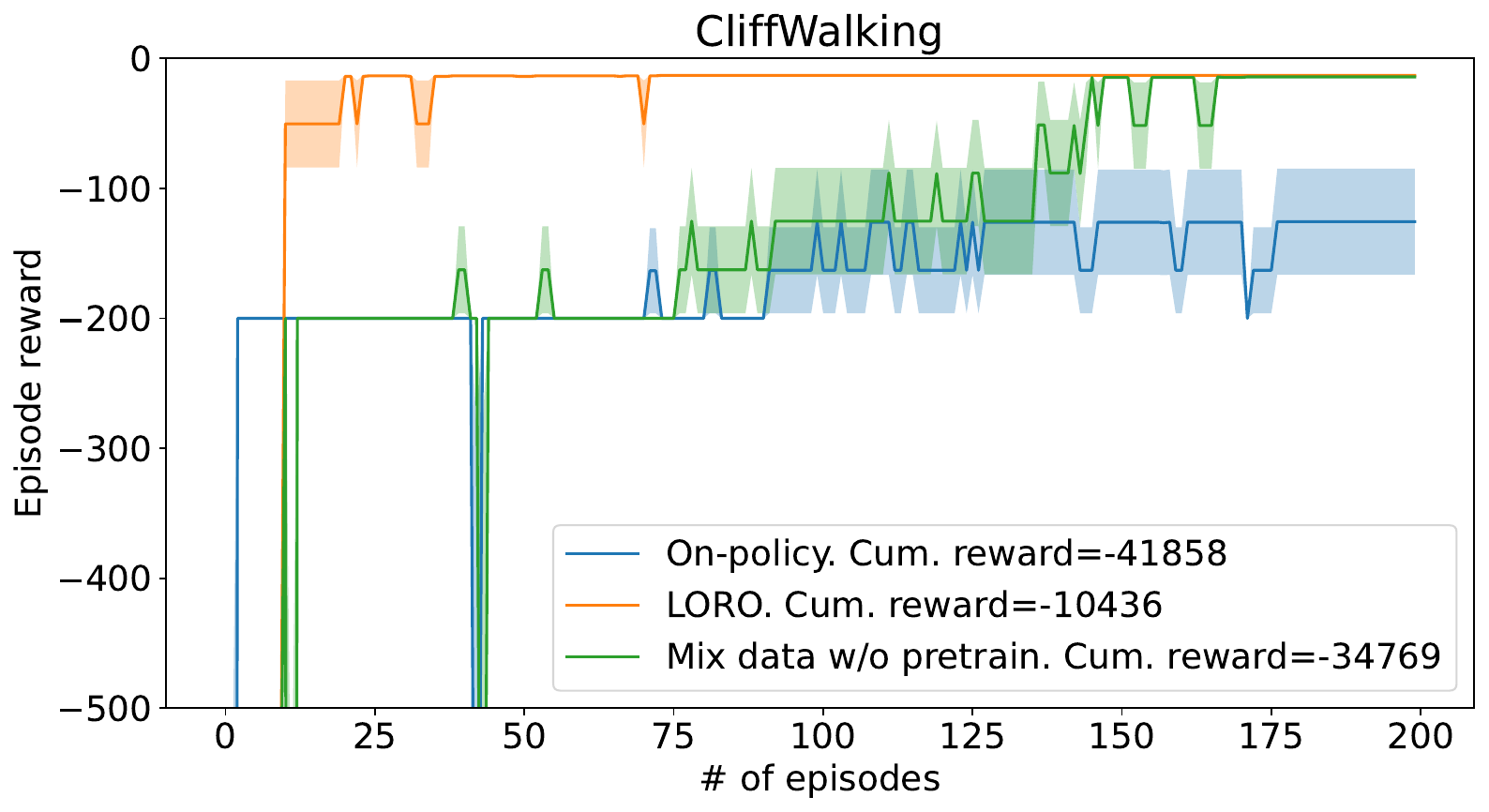}
\end{subfigure}\hspace{\fill} 
\begin{subfigure}[t]{0.48\textwidth}
    \includegraphics[width=\linewidth]{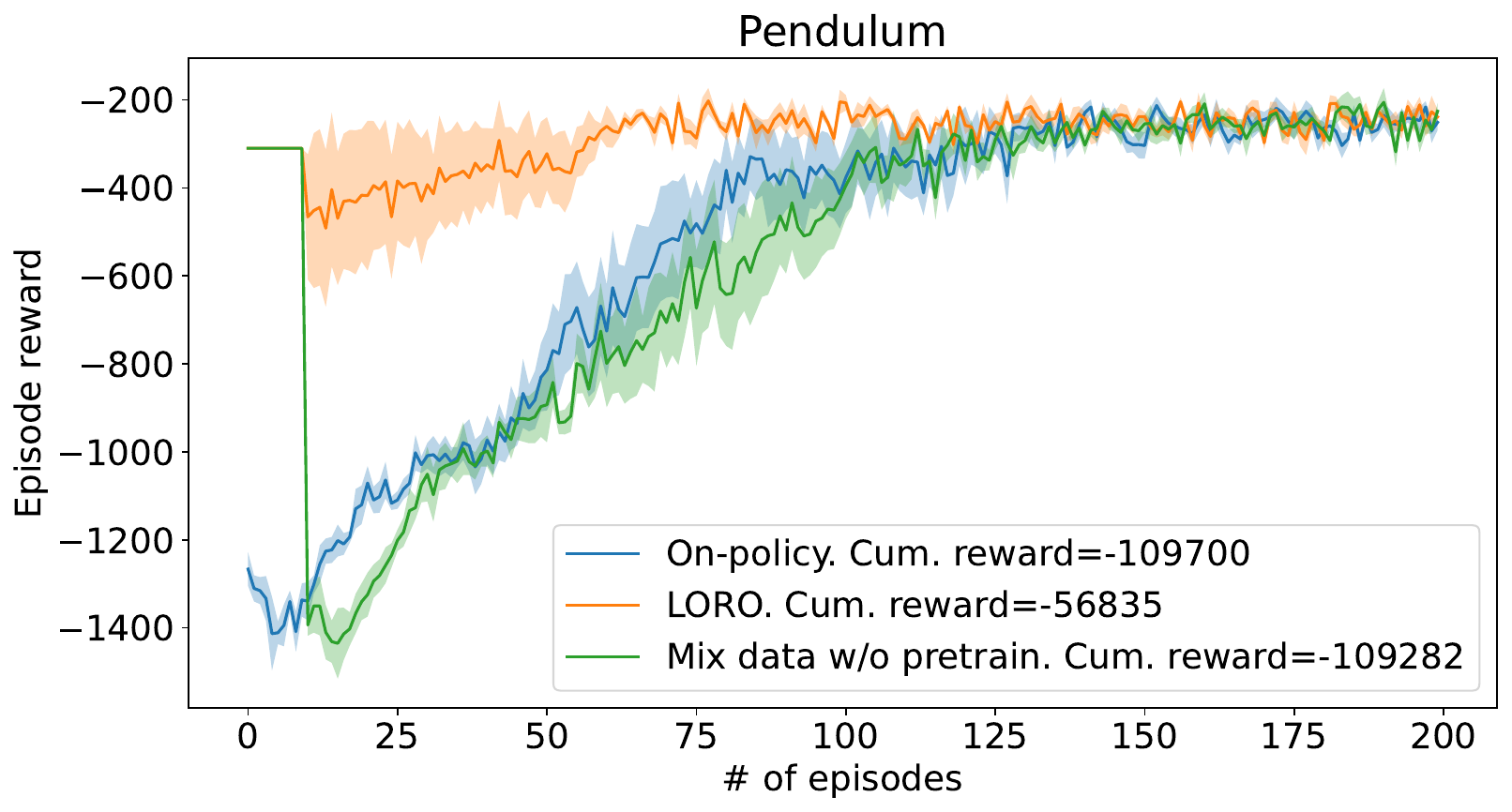}
\end{subfigure}

\begin{subfigure}[t]{0.48\textwidth}
    \includegraphics[width=\textwidth]{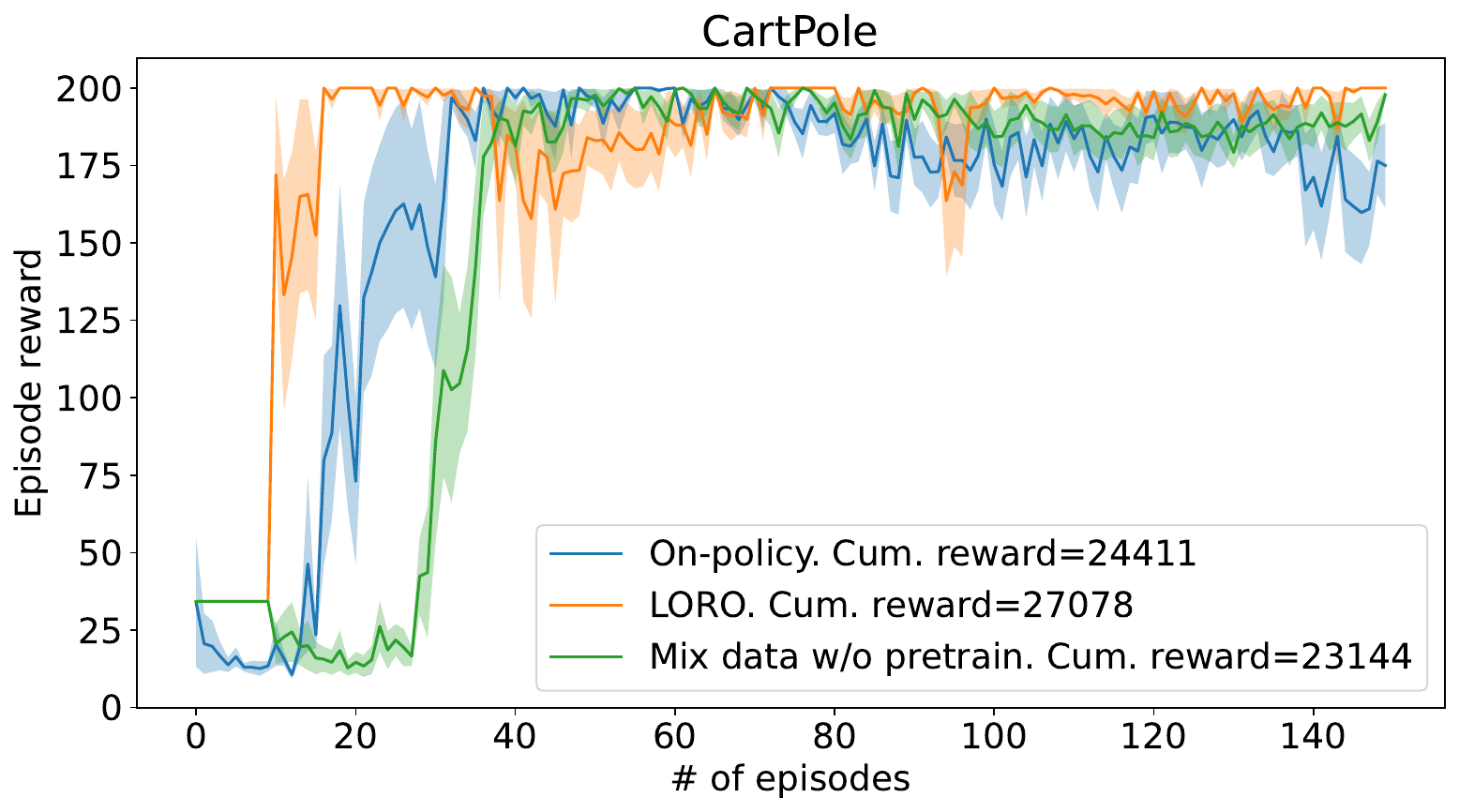}
\end{subfigure}\hspace{\fill} 
\begin{subfigure}[t]{0.48\textwidth}
    \includegraphics[width=\linewidth]{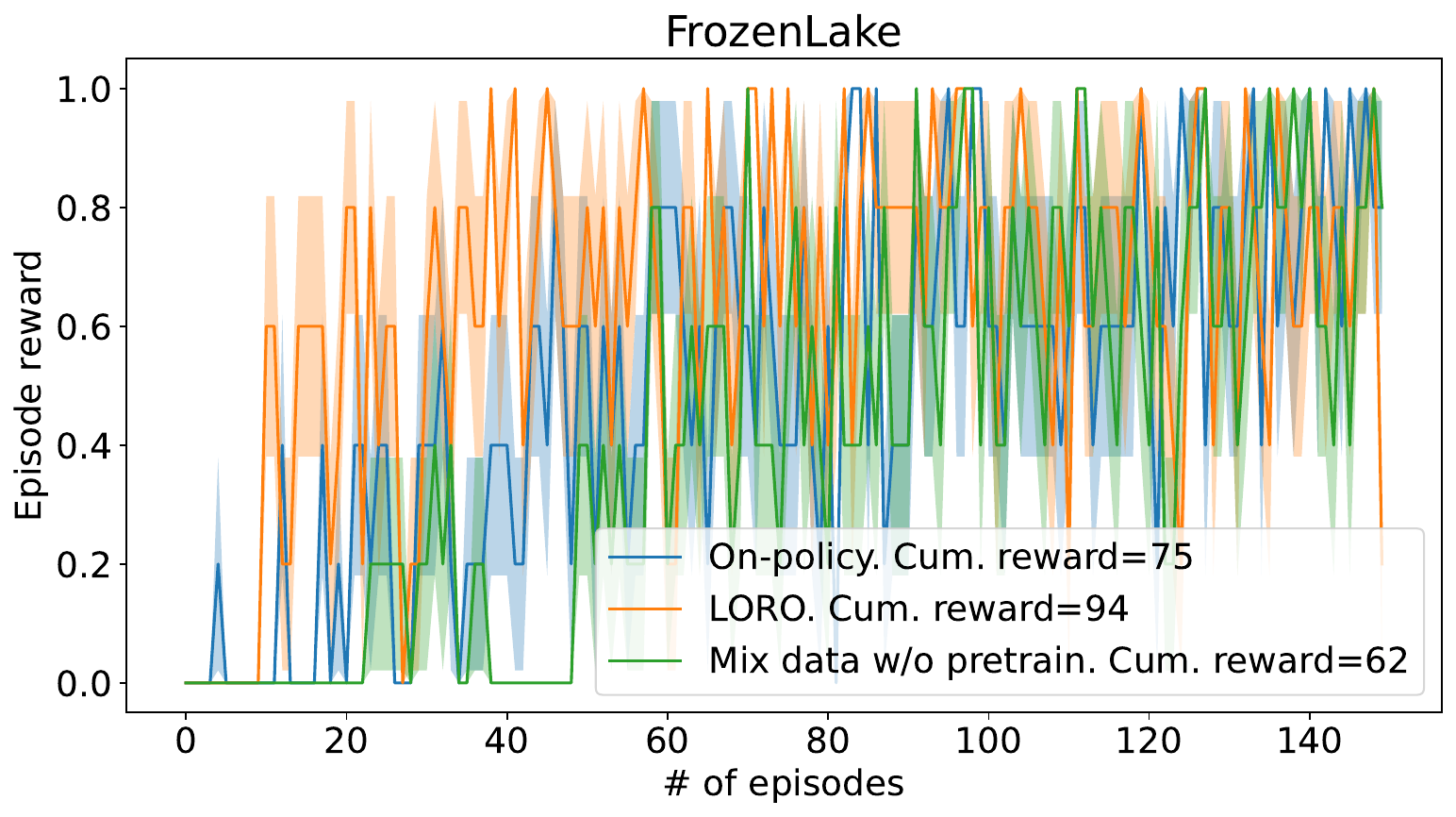}
\end{subfigure}

\begin{subfigure}[t]{0.48\textwidth}
    \includegraphics[width=\linewidth]{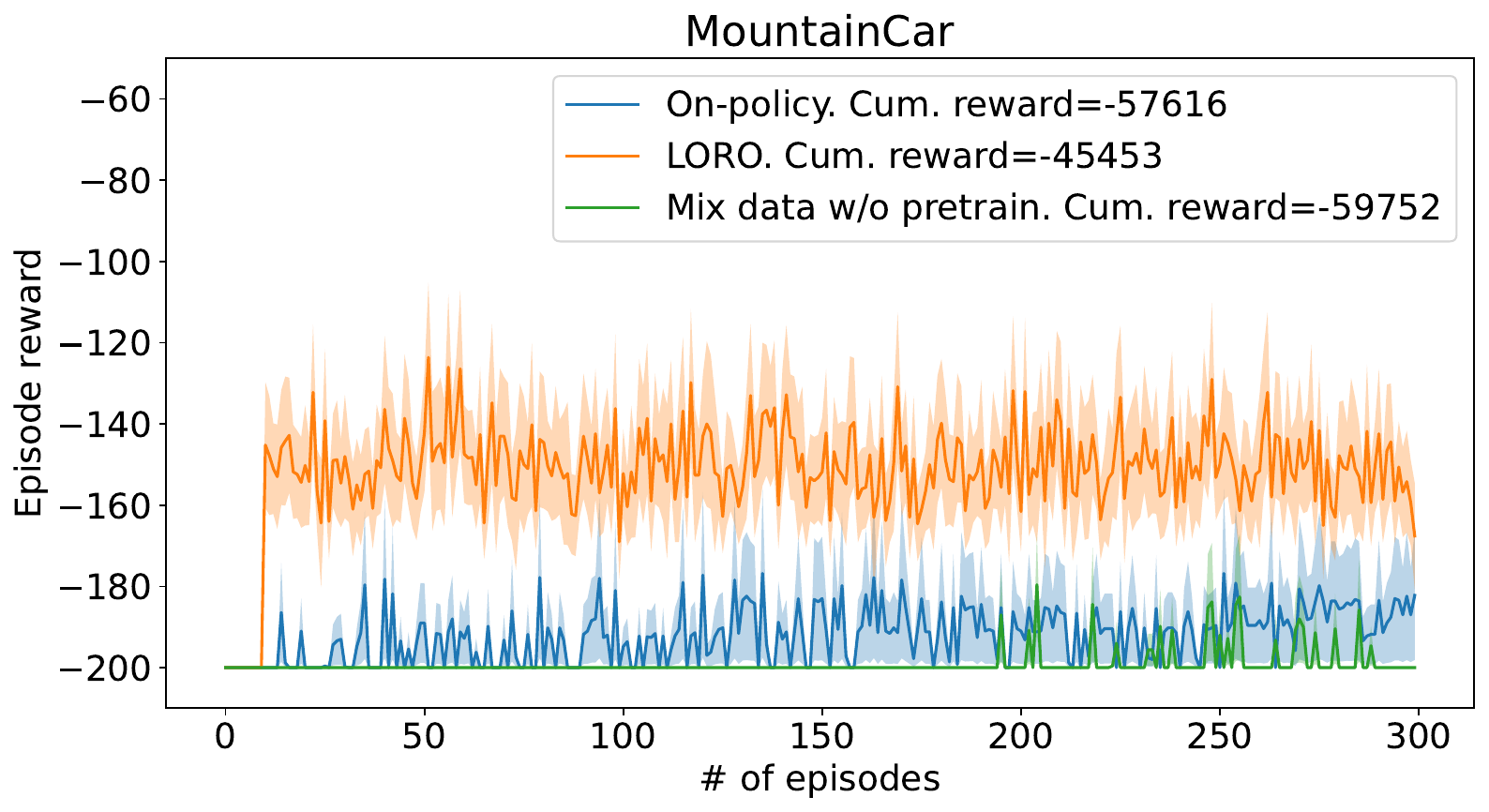}
\end{subfigure}\hspace{\fill} 
\begin{subfigure}[t]{0.48\textwidth}
    \includegraphics[width=\linewidth]{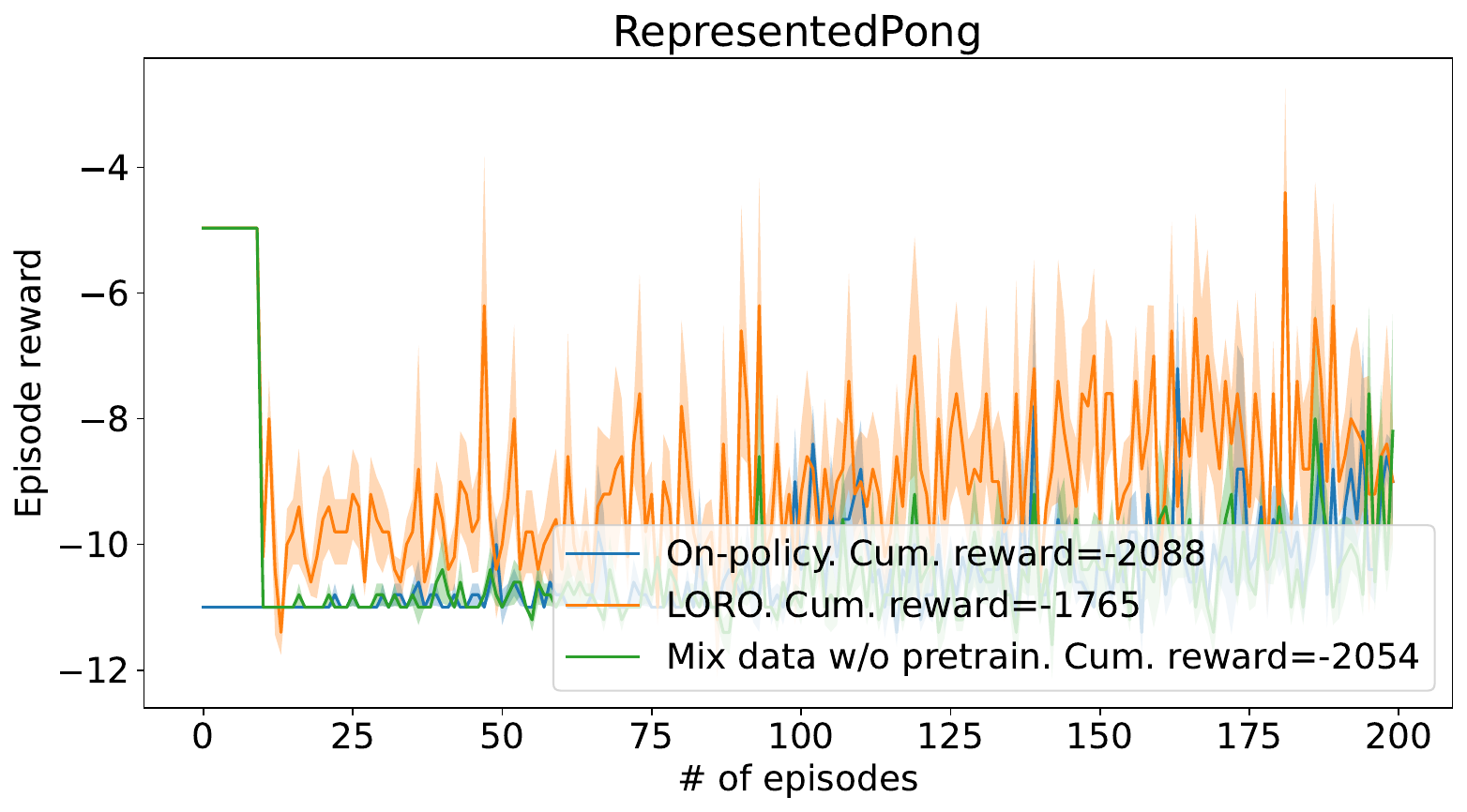}
\end{subfigure}
\caption{Comparing pre-training (then removing the collected data) versus mixing the LLM's collected data with on-policy data without pre-training.
It's clear that pre-training is necessary for LORO to achieve superior performance compared to naively mixing the data.}
\label{fig:pre_vs_mix}
\end{figure}

In Figure \ref{fig:pre_vs_mix}, we show that mixing the data alone (which is equivalent to \cite{song2022hybrid}) is insufficient. Our conjecture is that the pre-training step "overfits" LORO using only the high-quality data without the data from regions less visited by the optimal policy. As shown in Figure \ref{fig:main}, pre-training significantly boosts the performance of LORO compared to just mixing the LLM's collected data with the on-policy collected data after $\tau$ episodes.

Even though pre-training can be useful initially, to behave optimally, the agent still needs to explore other state-action pairs in case the initial data comes from a sub-optimal policy, as shown in the CartPole environment in Figure \ref{fig:main}.  




At first sight, there seems to be a contradiction between our findings and \cite{song2022hybrid}. ~\cite{song2022hybrid} assumes access to a large offline dataset and, along with \cite{nair2020awac}, wants to keep the policy less conservative toward the offline data by treating the online versus offline data equivalently. In contrast, we don't have access to offline data. We instead use LLM to collect a small number of high-quality data, thus, unlike \cite{song2022hybrid}, LORO puts more weight on these observations initially.
Our experiment showed that being conservative by ``overfitting'' to the LLM dataset can help learning more efficiently.



\subsection{Effect of LLM's data}
\label{subsec:effect_llmdata}

In the previous section, Figure \ref{fig:pre_vs_mix} shows us the importance of pre-training using data collected by LLM. 
In this section, we perform an ablation study that demonstrates that the quality of such data is crucial.

\begin{figure}[hbtp!]
\captionsetup[subfigure]{justification=Centering}

\begin{subfigure}[t]{0.48\textwidth}
    \includegraphics[width=\textwidth]{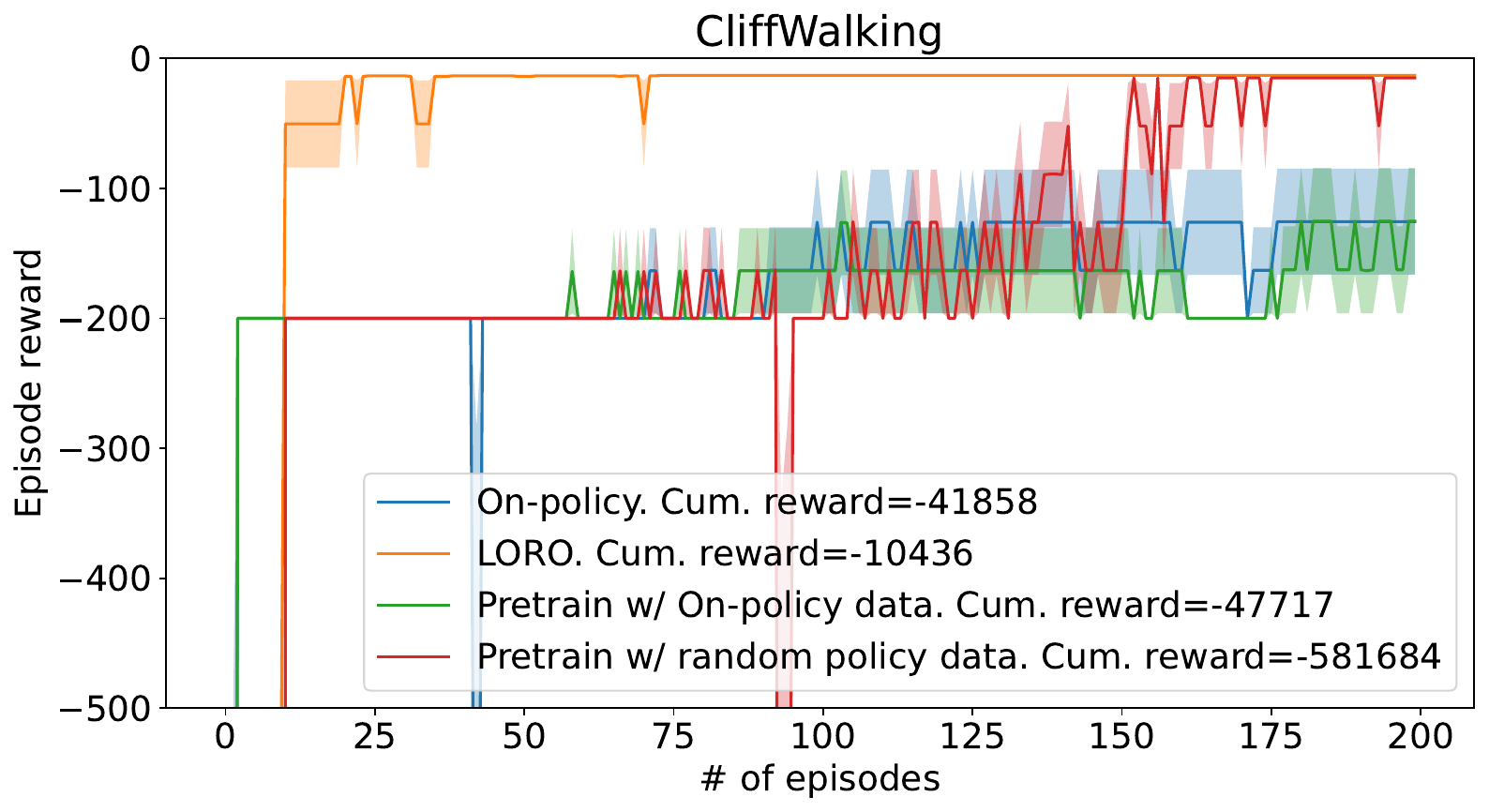}
\end{subfigure}\hspace{\fill} 
\begin{subfigure}[t]{0.48\textwidth}
    \includegraphics[width=\linewidth]{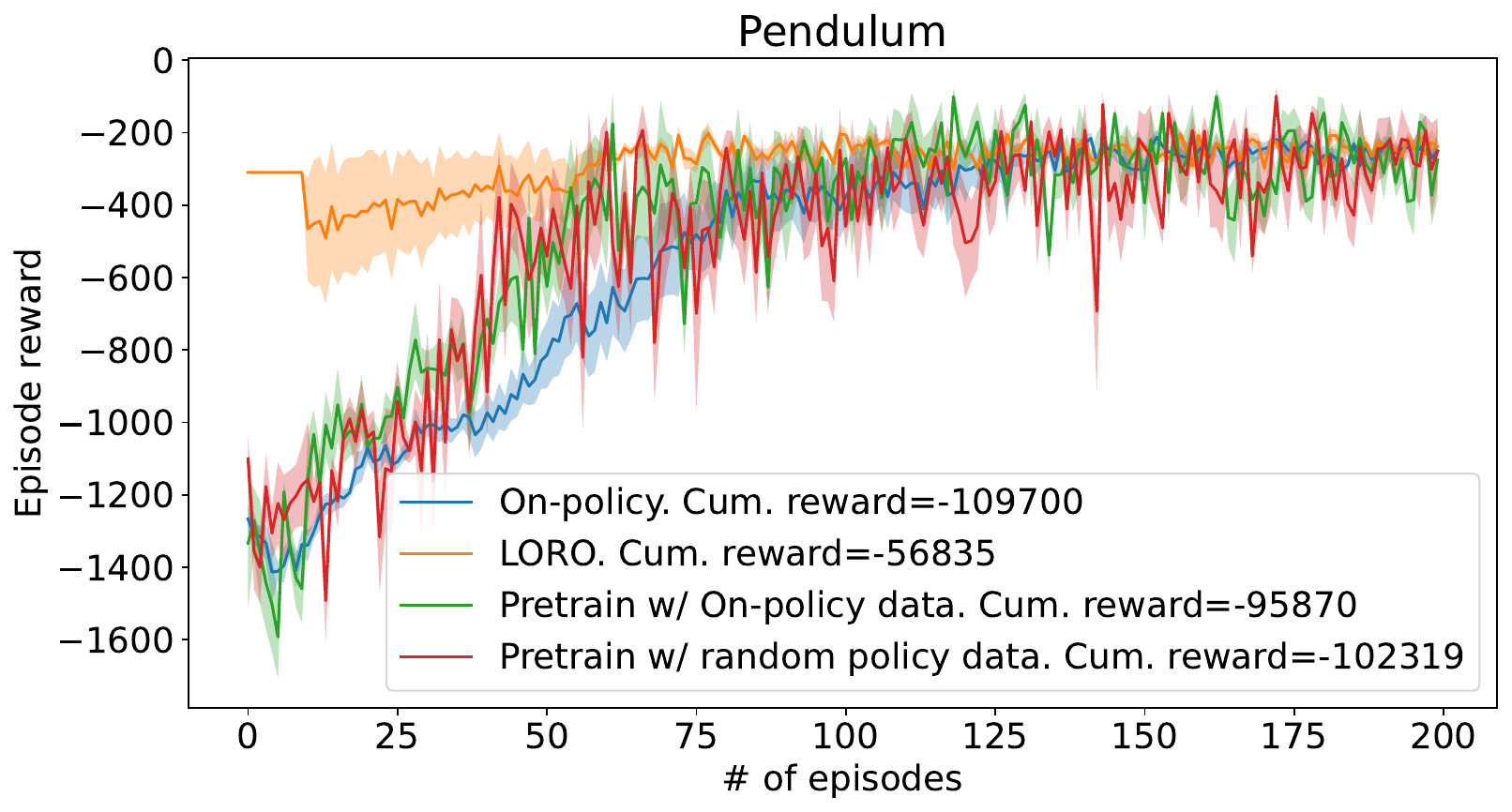}
\end{subfigure}

\begin{subfigure}[t]{0.48\textwidth}
    \includegraphics[width=\textwidth]{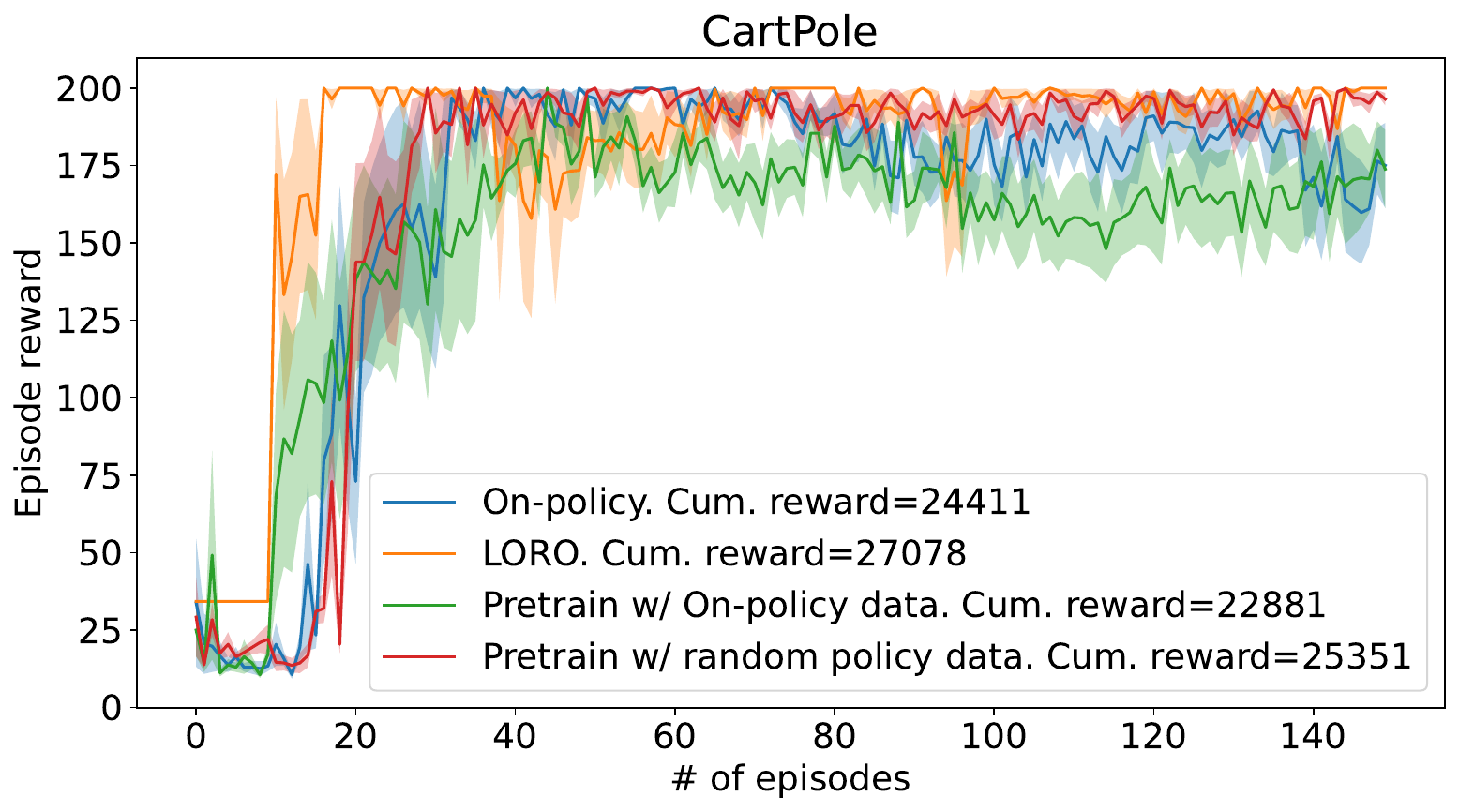}
\end{subfigure}\hspace{\fill} 
\begin{subfigure}[t]{0.48\textwidth}
    \includegraphics[width=\linewidth]{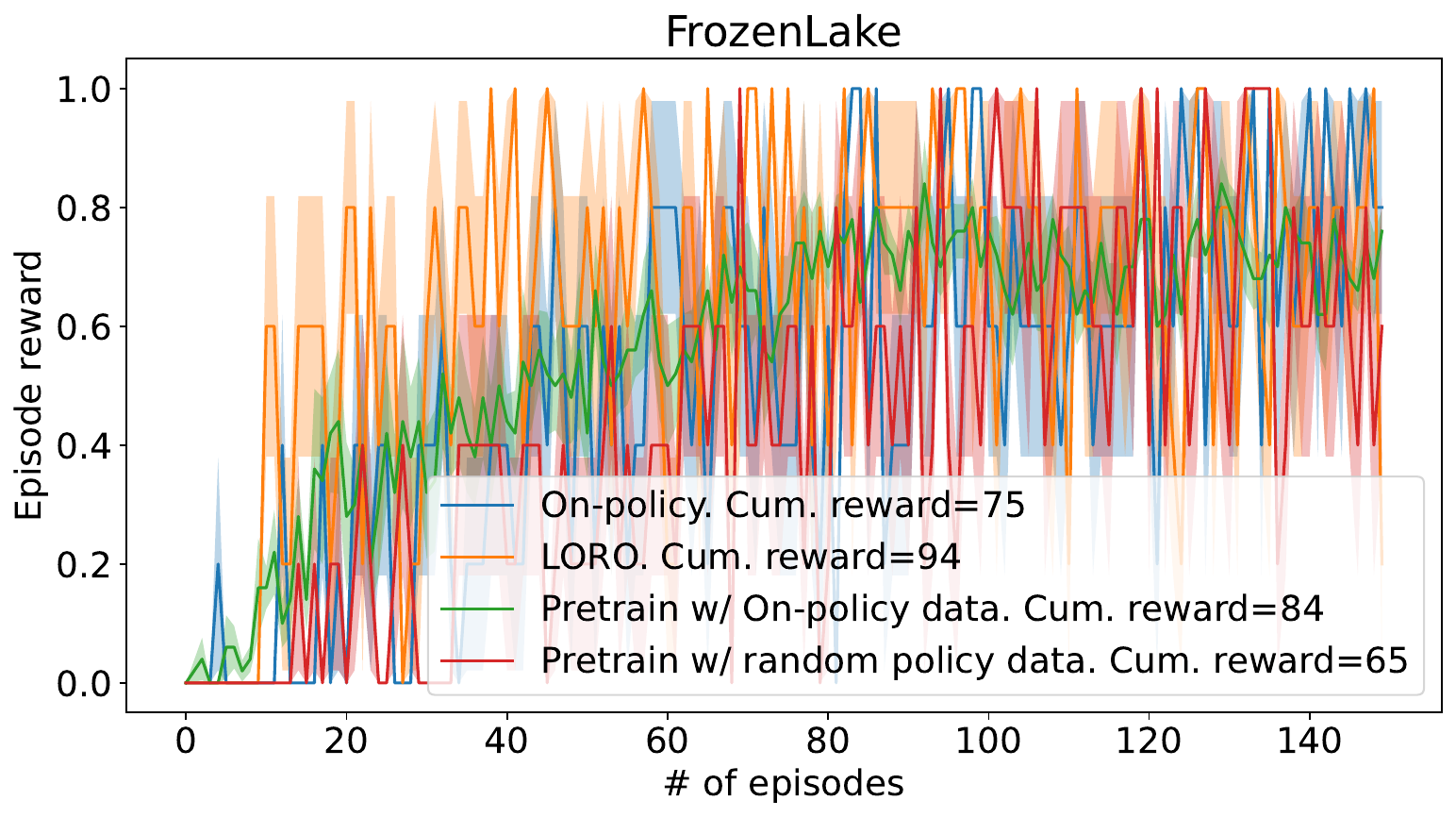}
\end{subfigure}

\begin{subfigure}[t]{0.48\textwidth}
    \includegraphics[width=\linewidth]{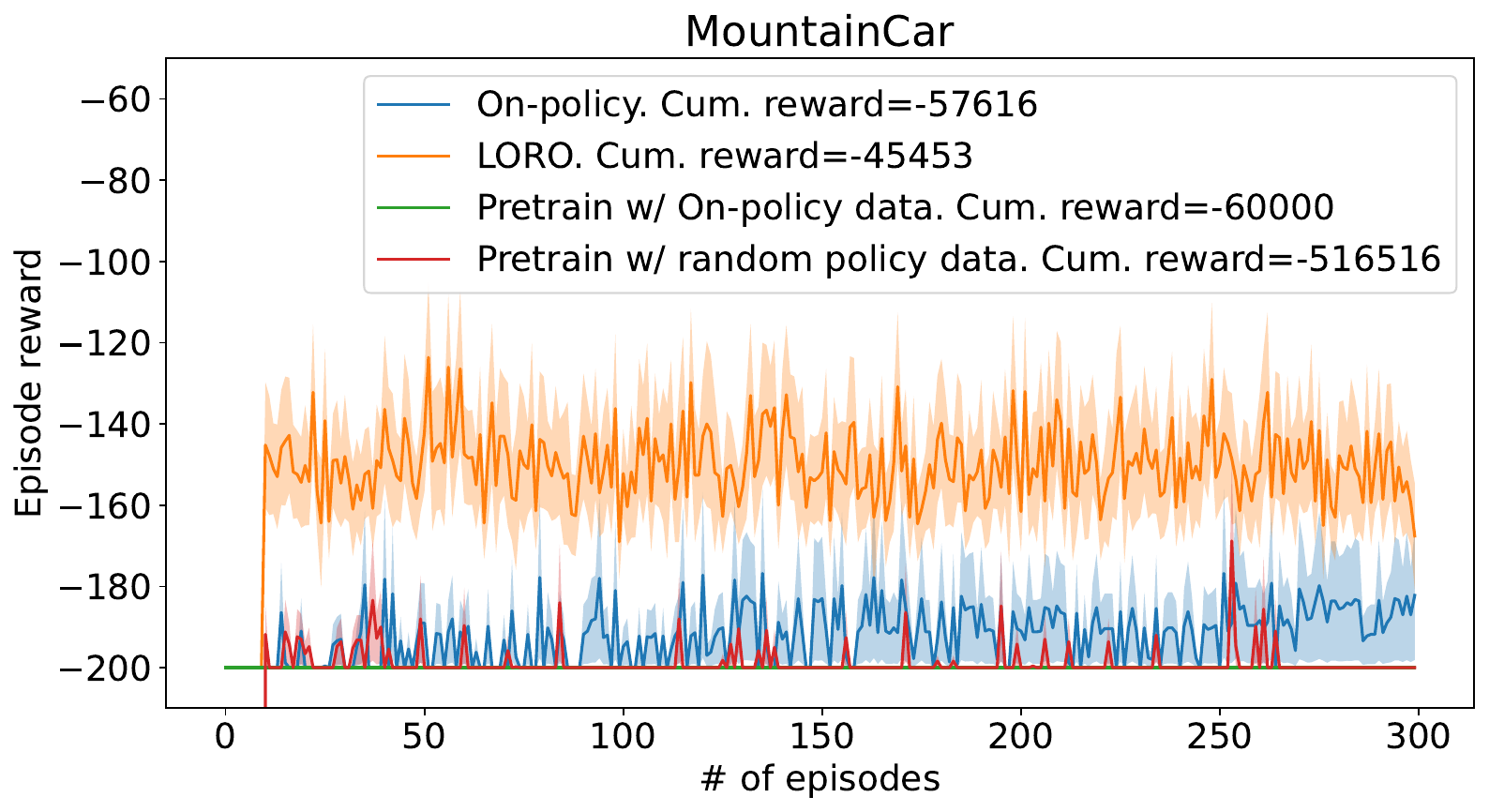}
\end{subfigure}\hspace{\fill} 
\begin{subfigure}[t]{0.48\textwidth}
    \includegraphics[width=\linewidth]{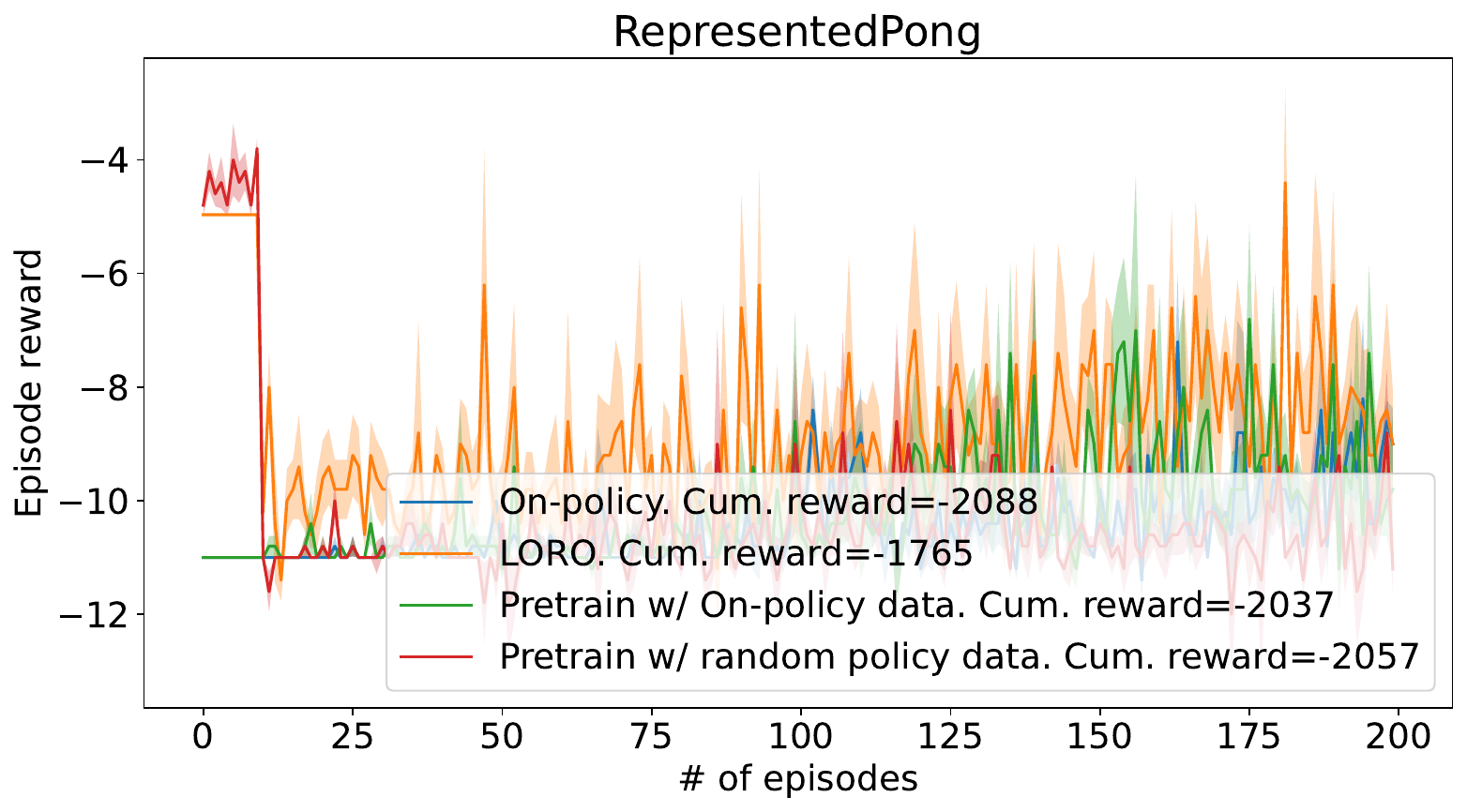}
\end{subfigure}
\caption{Comparing pre-training with LLM's data versus random and on-policy data. The main finding here is that pre-training is only useful with LLM's data.
}
\label{fig:llm_vs_other_data}
\end{figure}

In Figure \ref{fig:llm_vs_other_data}, we show that using LLM's collected data is significantly better than using data collected with an On-policy
RL algorithm from scratch
or a policy that takes actions uniformly at random. 
Thus, we conclude that pre-training is only beneficial when coupled with high-quality data, which supports our conjecture above.

\subsection{Effects of the LLM's capability}
\label{subsec:effect_llm_capability}

Given that the performance of many reasoning tasks increases with the improvement of the LLM's capability through: increasing the model's size, using Supervised Fine-Tuning (SFT), Long Chain-of-Thought (CoT), or some Test-time-scaling methods such as Majority Voting, and Best-of-N. In this section, we want to investigate whether this increase in LLM's reasoning capability also translates to decision making in MDP problems.

From our experiments in Appendix \ref{appendix:model_size}, we see no clear link between an LLM's model size and its performance. On the other hand, we notice that the LLM's base policies are only useful using CoT instead of just asking the LLM to make decisions. We also observe that the 0.5B model is not useful, as well as using Majority Voting or Best-of-N without CoT. Hence, in Appendix \ref{appendix_sft_lcot}, we want to see if increasing the LLM's capability using SFT or using an LLM with Long CoT can help.
%
We show that there is no significant difference in using standard CoT compared to using SFT or Long CoT. Understandably, SFT wouldn't be useful, or maybe even be counter-productive, since the amount of data collected for fine-tuning is too small to make a difference (around 500-3000 in our experiments). Hence, we conclude that improvements over LLM's capability does not directly translate to improvement in RL tasks.


\subsection{Other findings}

\begin{figure}[hbtp!]
    \includegraphics[width=\textwidth]{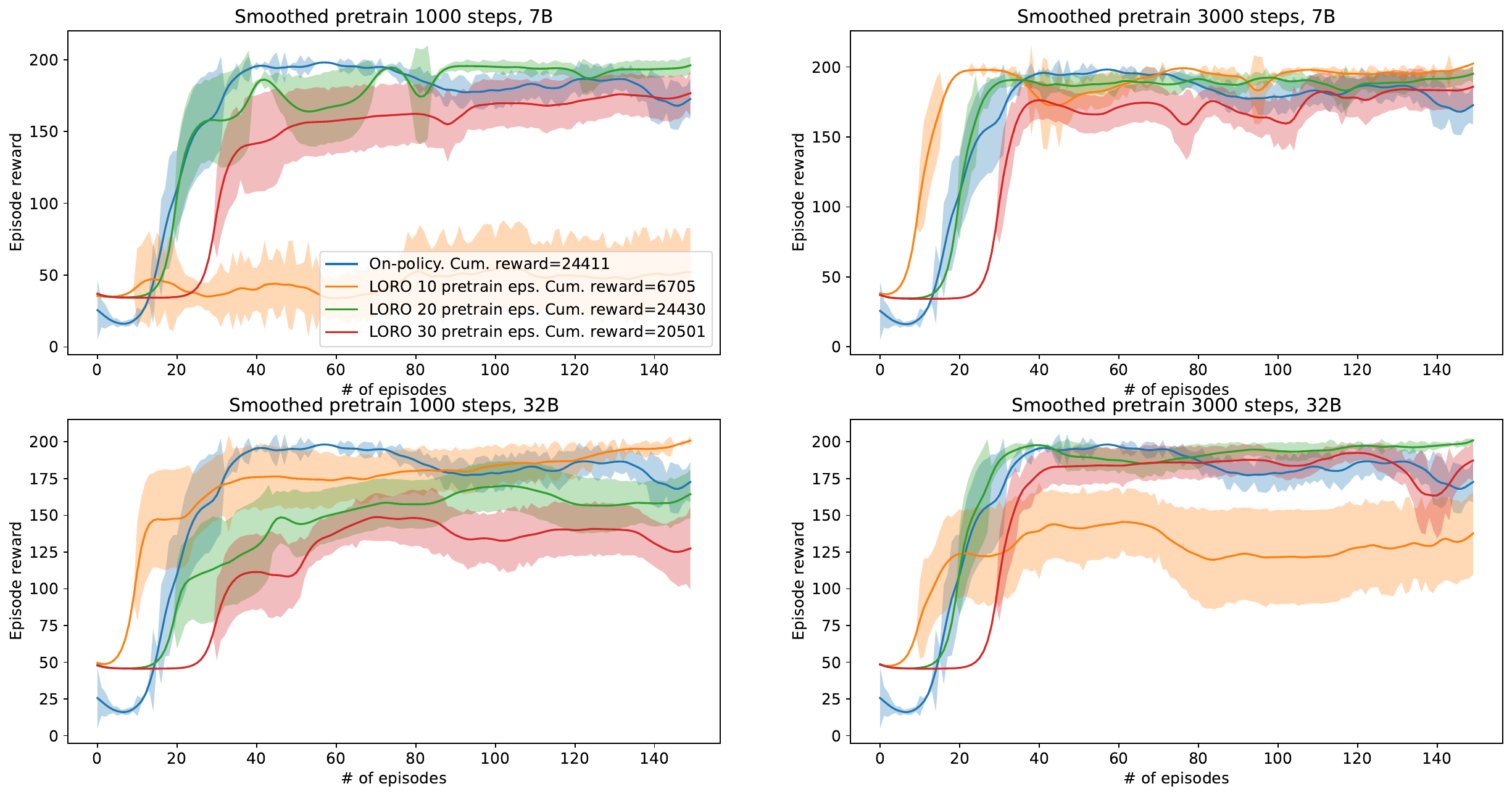}
    \caption{Comparing the effect of different amounts of pre-training data for the CartPole environment. This is a rare edge case where using LORO is worse than the On-policy baseline.}
    \label{fig:cartpole_bad}
\end{figure}

Interestingly, we find that a small model size (7B) is more sensitive to edge cases, such as MountainCar (with 3000 pre-training steps in Figure \ref{fig:mc_model_size}), CartPole (with 10 training episodes and 1000 pre-training steps in Figure \ref{fig:cp_model_size}), and FrozenLake (with bad environment history summarization in Figure \ref{fig:fl_bad_env_hist}). We also find that the amount of pre-training data in general does not affect
the cumulative reward, which was shown in Appendix \ref{appendix:pre_eps}. The only exception is the CartPole environment in Figure \ref{fig:cartpole_bad}, where it may be too easy for the On-policy baseline to learn the optimal policy.

Besides what we reported above, we find no clear relationship between the task's performance and the number of pre-training steps or the model size. These are shown in Appendix \ref{appendix:pre_step},  \ref{appendix:model_size}. We also found no clear difference between environments with Discrete Action versus Continuous Action (e.g., Pendulum), despite the intuition that the Discrete Action environments should be easier for the LLM \cite{singh2025optionspitfallsmultiplechoicequestions}.  

\section{Conclusion and Future work}

In this paper, we investigate how to leverage an LLM to warm-start traditional RL methods. Empirically, we have shown that the high-quality data collected by the LLM, 
can significantly increase the sample efficiency of online RL. Our algorithm, under Assumption \ref{assum:coverage}, is supported by the Cumulative suboptimality regret and Sample complexity guarantee from \cite{song2022hybrid}.

Our work provides a framework for significantly reducing the sample complexity in RL problems. This is especially important for practical applications where the data collection cost or safety is a major concern. A limitation of our work is that Assumption \ref{assum:coverage} may not hold for some RL tasks, but we believe that the increasing capability of LLM would increase the range of problems where Assumption \ref{assum:coverage} is applicable. In the future, we would like to extend this work to more sophisticated RL problems, with a large State and Action space. We would also like to investigate how to scale the sample efficiency with the LLM's capability.


\bibliographystyle{plainnat}
\bibliography{learning}

\newpage
\appendix

\section{Ablation study}

\subsection{The environments and implementation details}
\label{sec:appendix_env_impl}
\subsubsection{The environments}
We empirically verify our algorithm on some classic RL environments:

\begin{itemize}
    \item \textbf{Cart Pole}: The agent aims to balance a pole on top of a cart by moving left and right. It observes the Cart Position, Cart Velocity, Pole Angle, and Pole Angular Velocity. The reward is one for every step taken before the episode ends, either by having the pole fall over, moving the cart to the edge of the display, or reaching the maximum episode length.
    \item \textbf{Pendulum}: The agent aims to swing up an inverted pendulum by applying torque on its free end. It observes the $(x,y)$ location of the pendulum's free end and its angular velocity. From the location, we calculate the pendulum's angle and the rotating direction to help the LLM, but do not use them in the online phase. The reward is calculated based on the pendulum angle, where the upright location has the highest reward. The episode ends when it reaches the maximum episode length. Note that the action set here is continuous, which can be more challenging for the LLM's policy.
    \item \textbf{Frozen Lake}: The agent aims to move from the top-left to the bottom-right location in a four-by-four grid world. The agent can move up, down, left, and right. It only observes its own location. The reward is zero everywhere except at the goal, where the reward is one. The episode ends either when the agent moves to one of the four "holes" in the grid, reaches the goal, or reaches maximum episode length. We further implement an external environment history to store the rewards received at each visited location, which is necessary for the LLM to solve this task. The environment's history is not used in the online learning phase.
    \item \textbf{Cliff Walking}: The agent aims to move from the bottom-left to the bottom-right location in a four-by-twelve grid world. The agent can move up, down, left, and right. It only observes its own location. The reward is negative one everywhere except negative one hundred at the cliff locations on the bottom of the grid. The episode ends either when the agent reaches the goal or reaches maximum episode length. We also use the environment history for this environment.
    \item \textbf{Represented Pong}: This is the Atari game Pong, but instead of the traditional image observation, we use \cite{anand2019unsupervised} to extract the game state information from the RAM state. The agent then observes its own coordination, the ball's, the opponent's, and the score. We also calculate the ball velocity and add it to the observation, since it seems necessary to ensure Markov's property (able to take optimal action with only the current state information). The agent controls the right paddle up and down and competes against the left paddle controlled by the computer by trying to deflect the ball away from your goal and into the opponent's goal. The agent receives a point whenever it scores a goal and loses when the opponent does. The game ends when a player's score reaches twenty-one or the agent reaches the maximum episode length.
    \item \textbf{Mountain Car}: The agent's goal is to move from the bottom of a sinusoidal valley to the top of the right hill as quickly as possible. The agent can strategically accelerate left or right. It only observes its location and velocity. The reward is negative one everywhere except the goal. The episode ends either when the agent reaches the goal or reaches maximum episode length.
\end{itemize}

\subsubsection{Implementation details}

We build our code from \cite{pandey2024llamagym}, which provides a framework for LLM interacting with OpenAI's gym games with a built-in text description wrapper to turn RL games into something LLM can play. The game descriptions, which are listed in Appendix \ref{appendix:prompt}, are heavily referenced from \cite{text2024gymagent}. The RL training process is using d3rlpy \cite{d3rlpy}, with the default hyperparameter choice, with batch-size 256, buffer size 100,000, $\varepsilon:$ 0.1, $\gamma:$ 0.99, target update interval 1,000, and learning rate $5e-5$. We use DDQN \cite{vanhasselt2015deepreinforcementlearningdouble} for all tasks with Discrete Action and SAC \cite{haarnoja2019softactorcriticalgorithmsapplications} for Continuous Action. The LLM was run on two H100 GPUs.

For the LORO algorithm, we collected data to pre-train a policy and then only used online data in the online learning process.

\subsection{Effects of LLM's model size}
\label{appendix:model_size}

\begin{figure}[hbtp!]
    \includegraphics[width=\textwidth]{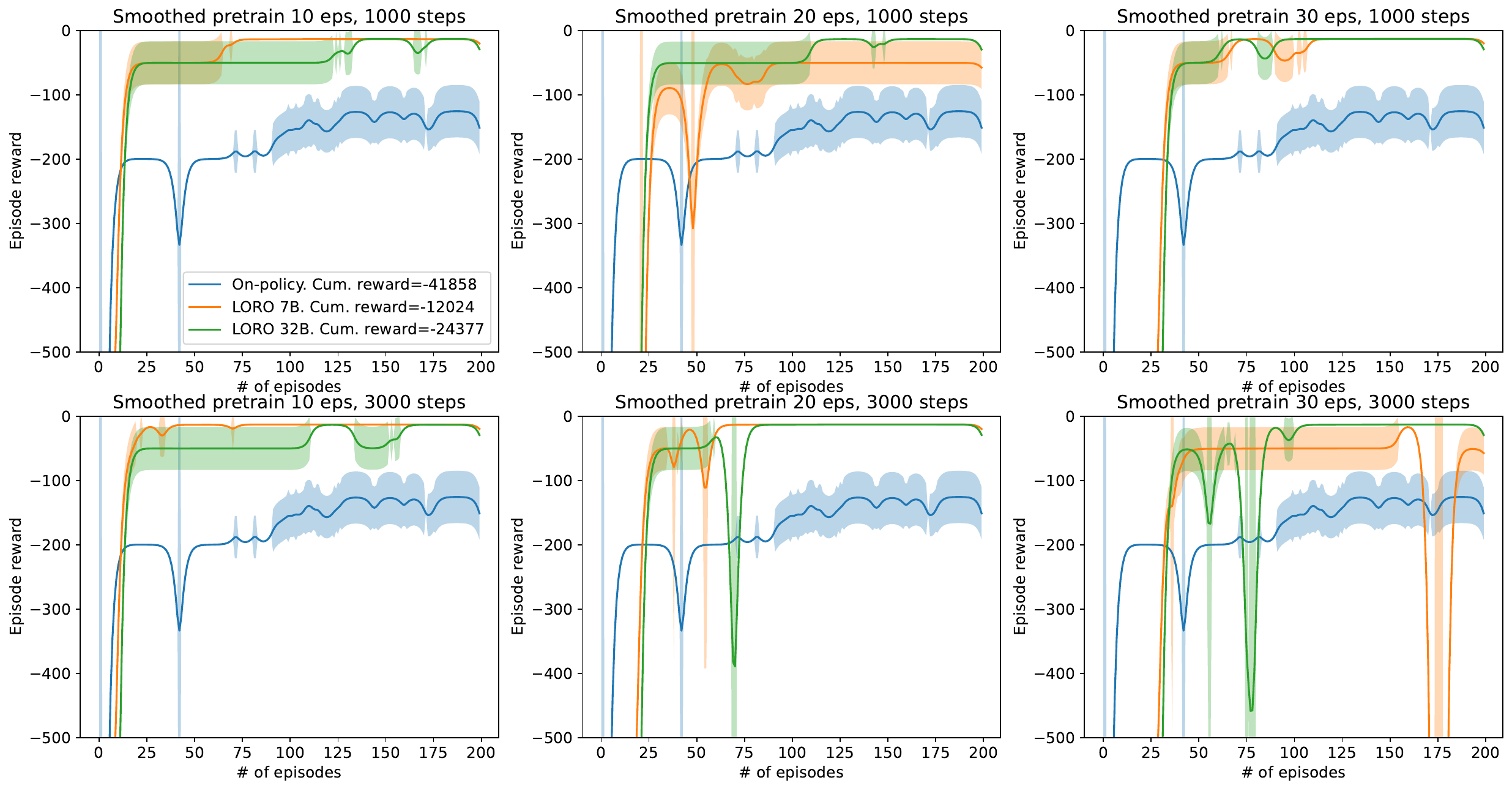}
    \caption{Comparing the effect of different LLMs' model sizes for the CliffWalking environment.}
\label{fig:cw_model_size}
\end{figure}

\begin{figure}[hbtp!]
    \includegraphics[width=\linewidth]{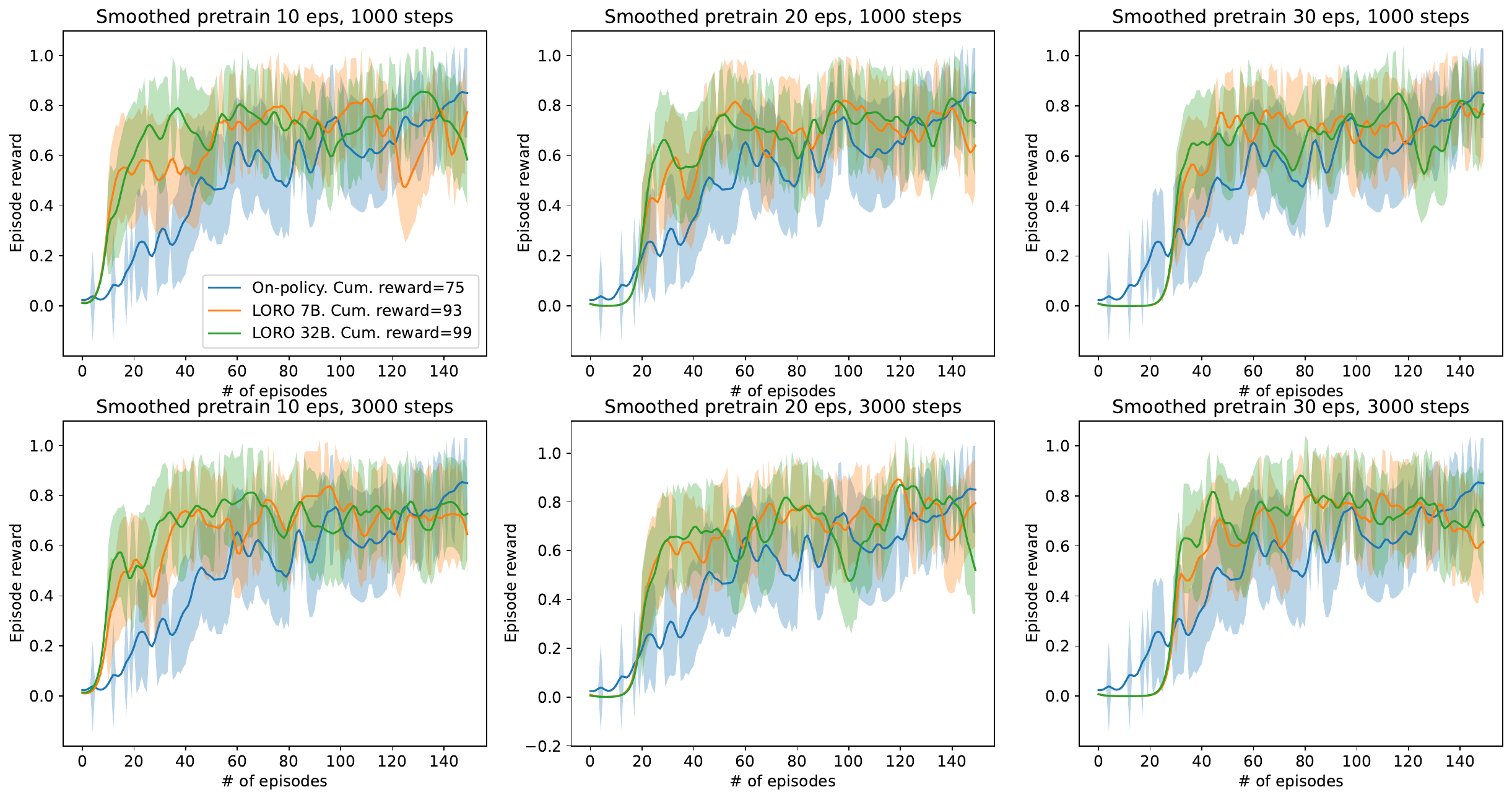}
    \caption{Comparing the effect of different LLMs' model sizes for the FrozenLake environment.}
\label{fig:fl_model_size}
\end{figure}

\begin{figure}[hbtp!]
    \includegraphics[width=\textwidth]{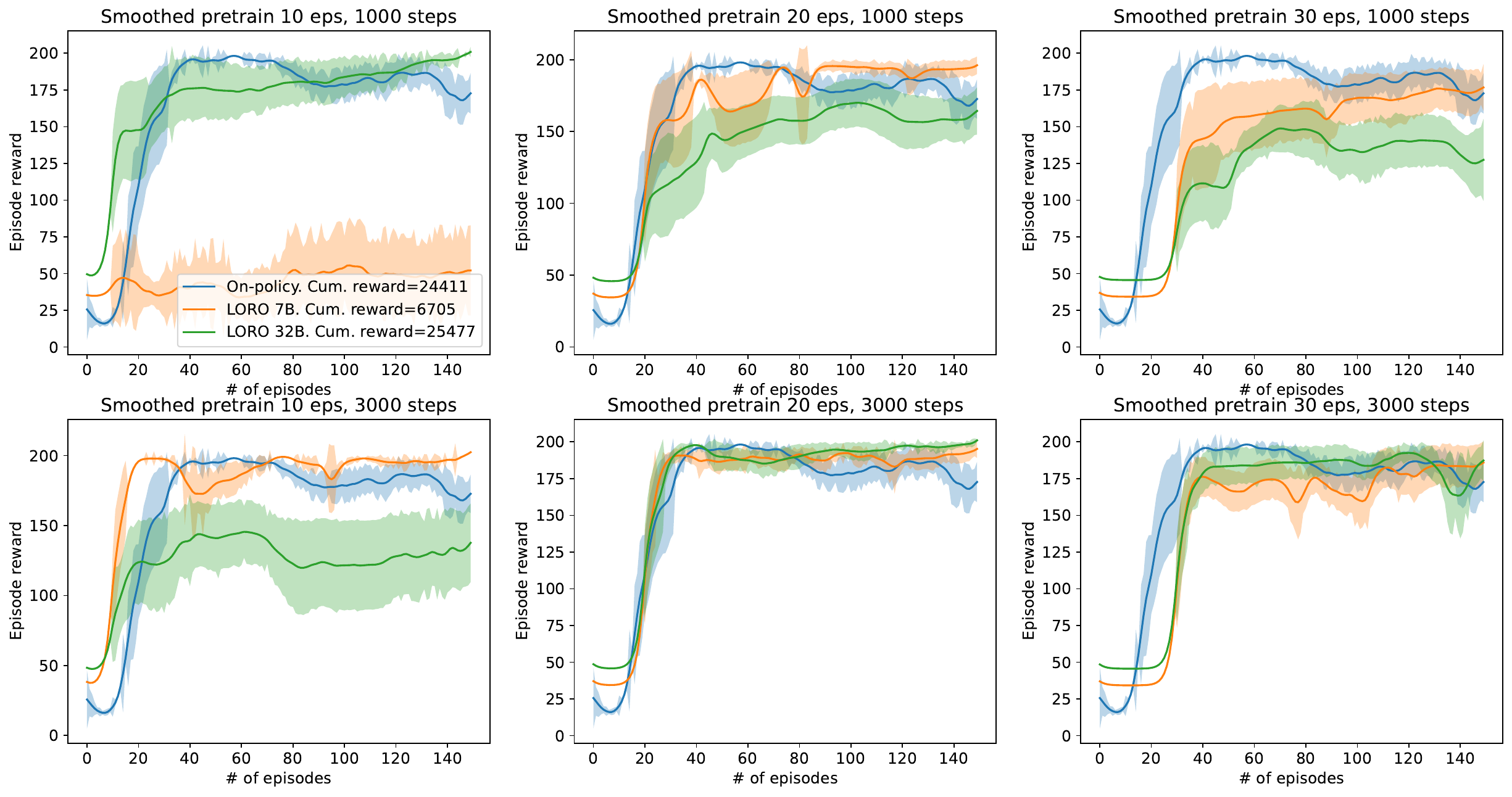}
    \caption{Comparing the effect of different LLMs' model sizes for the CartPole environment.}
\label{fig:cp_model_size}
\end{figure}

\begin{figure}[hbtp!]
    \includegraphics[width=\linewidth]{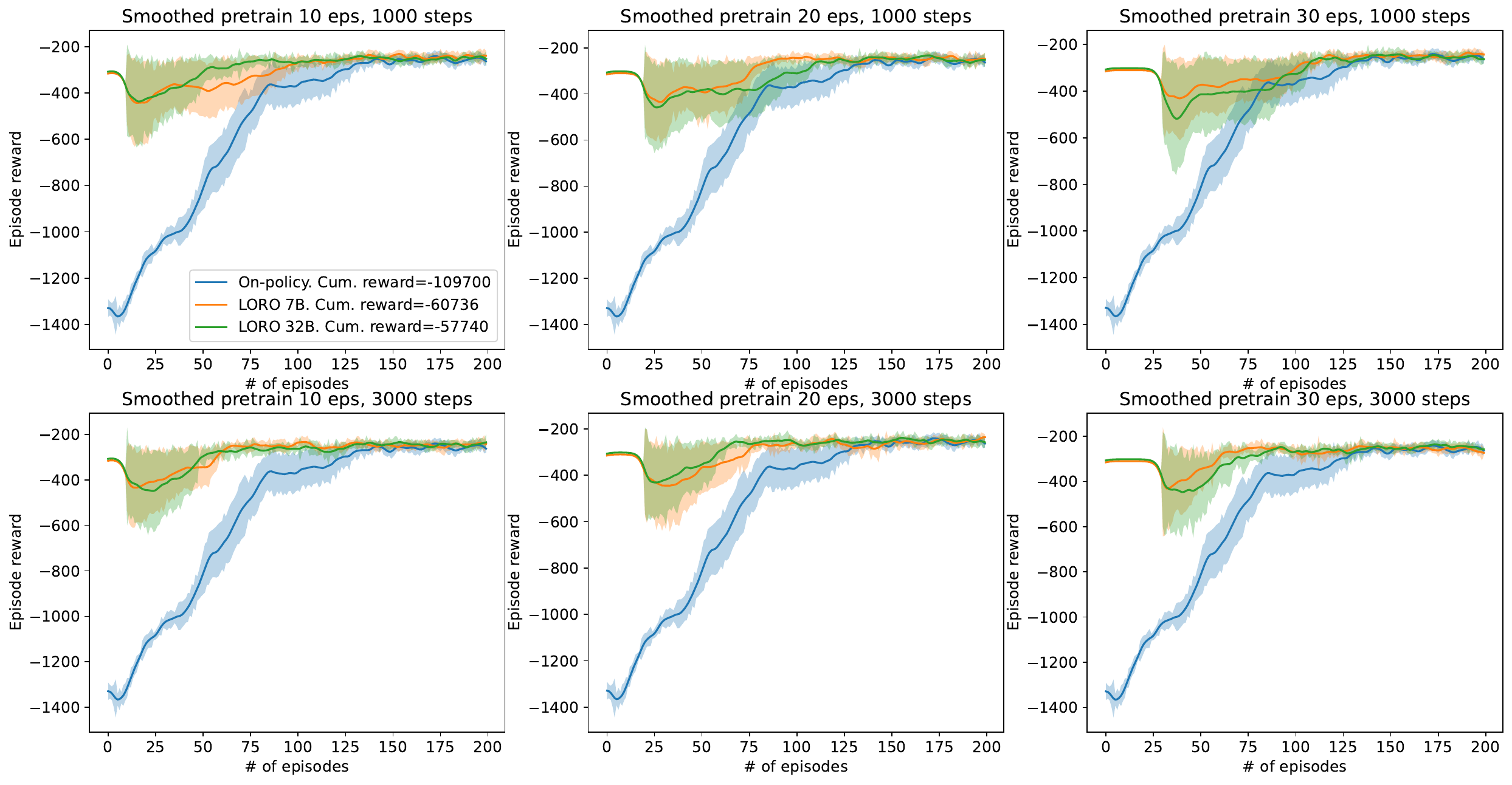}
    \caption{Comparing the effect of different LLMs' model sizes for the Pendulum environment.}
\label{fig:pe_model_size}
\end{figure}

\begin{figure}[hbtp!]
    \includegraphics[width=\linewidth]{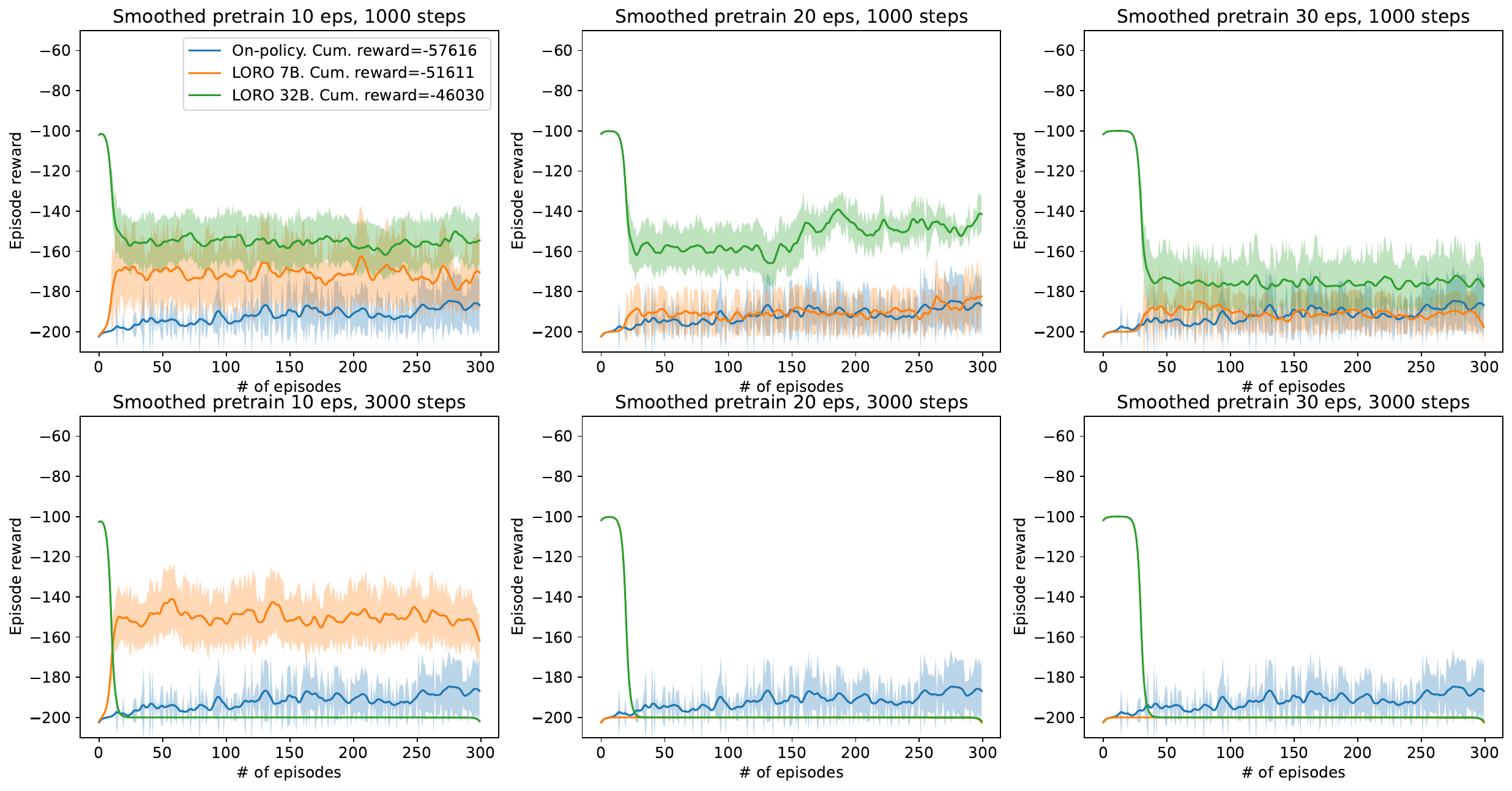}
    \caption{Comparing the effect of different LLMs' model sizes for the MountainCar environment.}
\label{fig:mc_model_size}
\end{figure}

\begin{figure}[hbtp!]
    \includegraphics[width=\linewidth]{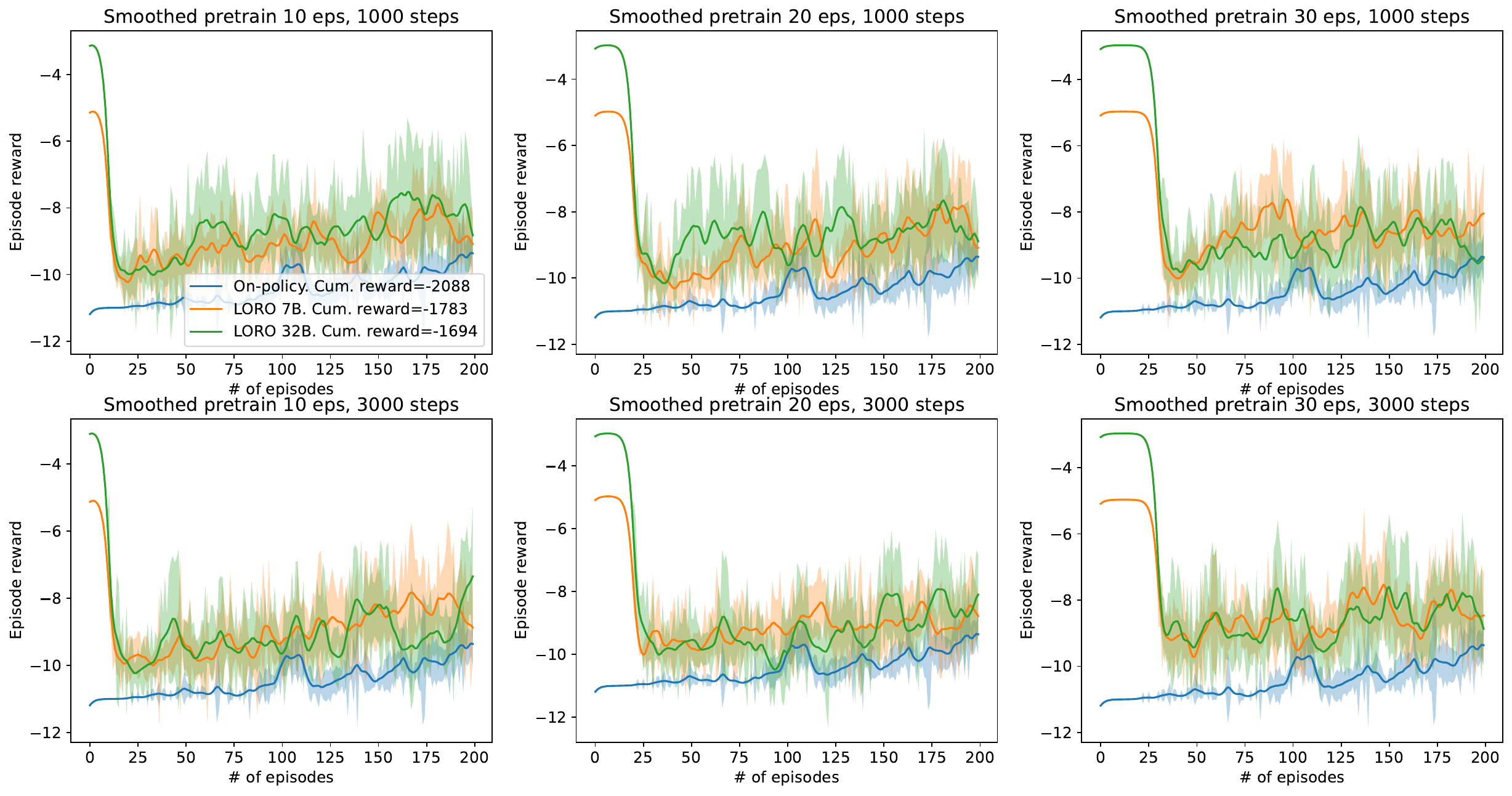}
    \caption{Comparing the effect of different LLMs' model sizes for the Pong environment.}
\label{fig:rp_model_size}
\end{figure}

In this section, we evaluate the effect of the LLM's model size on the cumulative reward of the policy. We evaluate this with different pre-training data and pre-training steps on six OpenAI Gym environments and show the result in Figure \ref{fig:cw_model_size}, \ref{fig:fl_model_size}, \ref{fig:cp_model_size}, \ref{fig:pe_model_size}, \ref{fig:mc_model_size}, \ref{fig:rp_model_size}. Overall, we observe no clear advantage of using a larger model to improve the decision-making quality of the LORO policy.





\newpage
\subsection{Effects of the number of pre-training steps}
\label{appendix:pre_step}

\begin{figure}[hbtp!]
    \includegraphics[width=\textwidth]{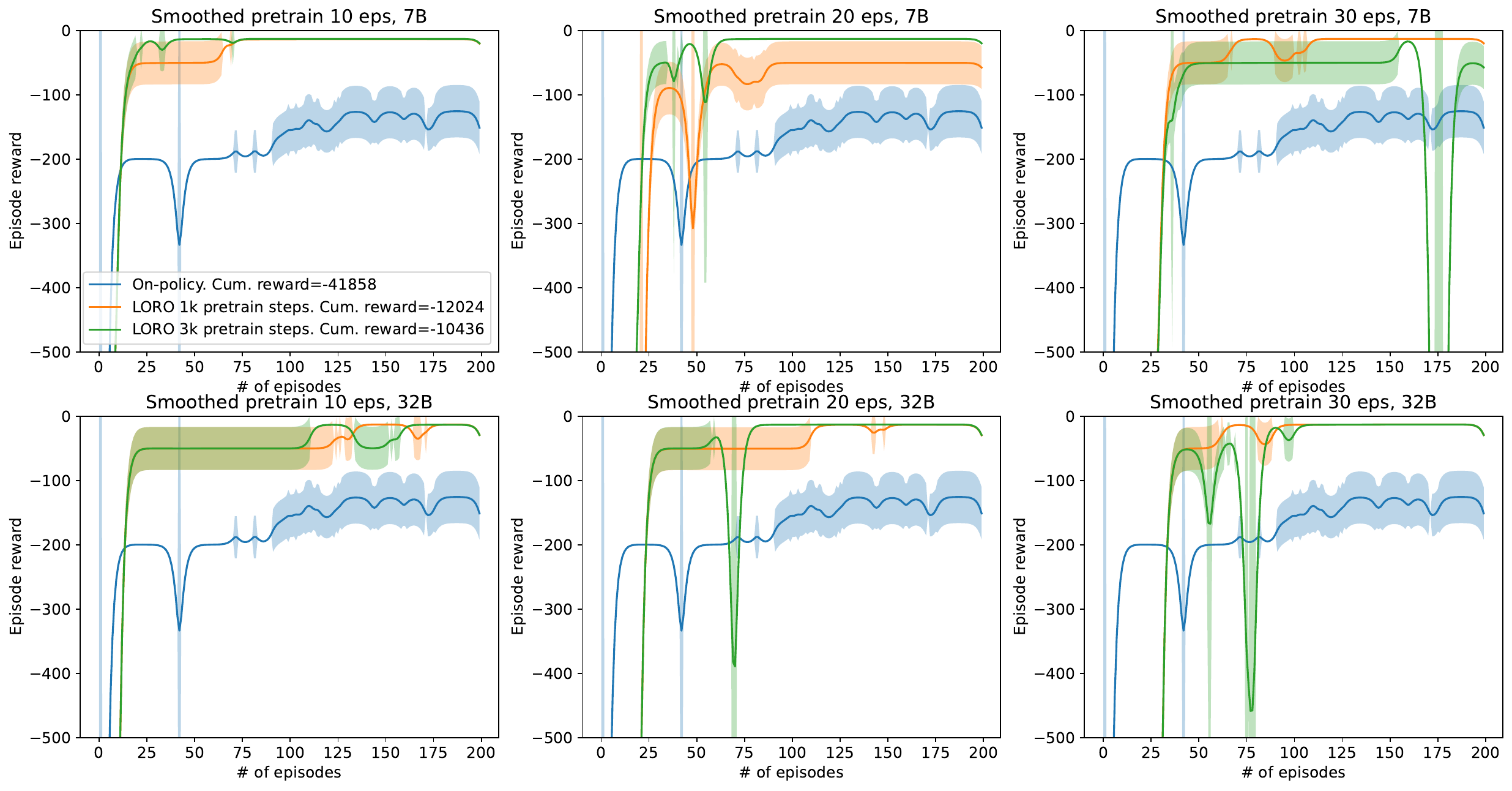}
    \caption{Comparing the effect of different pre-training steps for the CliffWalking environment.}
\label{fig:cw_pre_steps}
\end{figure}

\begin{figure}[hbtp!]
    \includegraphics[width=\linewidth]{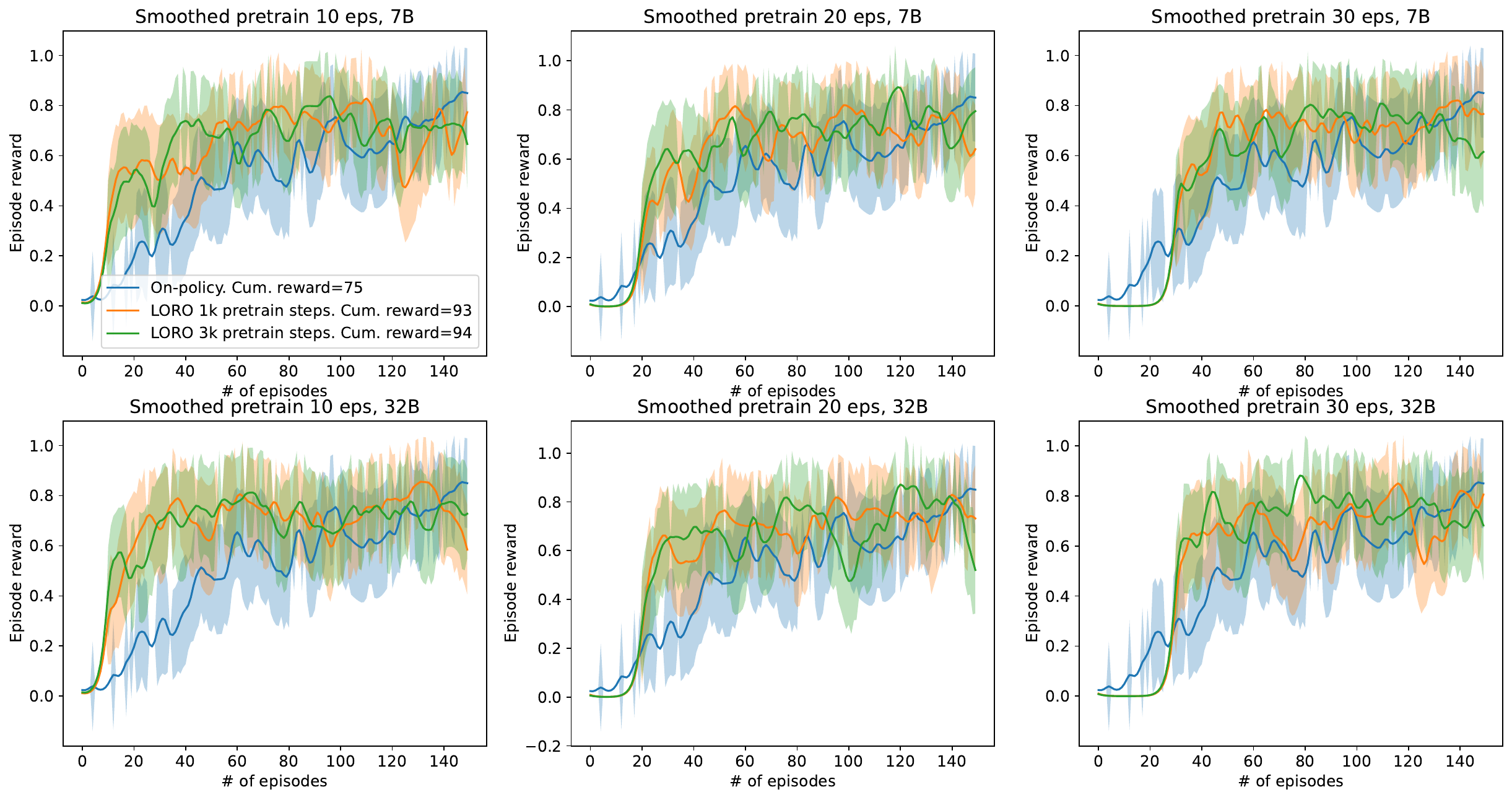}
    \caption{Comparing the effect of different pre-training steps for the FrozenLake environment.}
\label{fig:fl_pre_steps}
\end{figure}

\begin{figure}[hbtp!]
    \includegraphics[width=\textwidth]{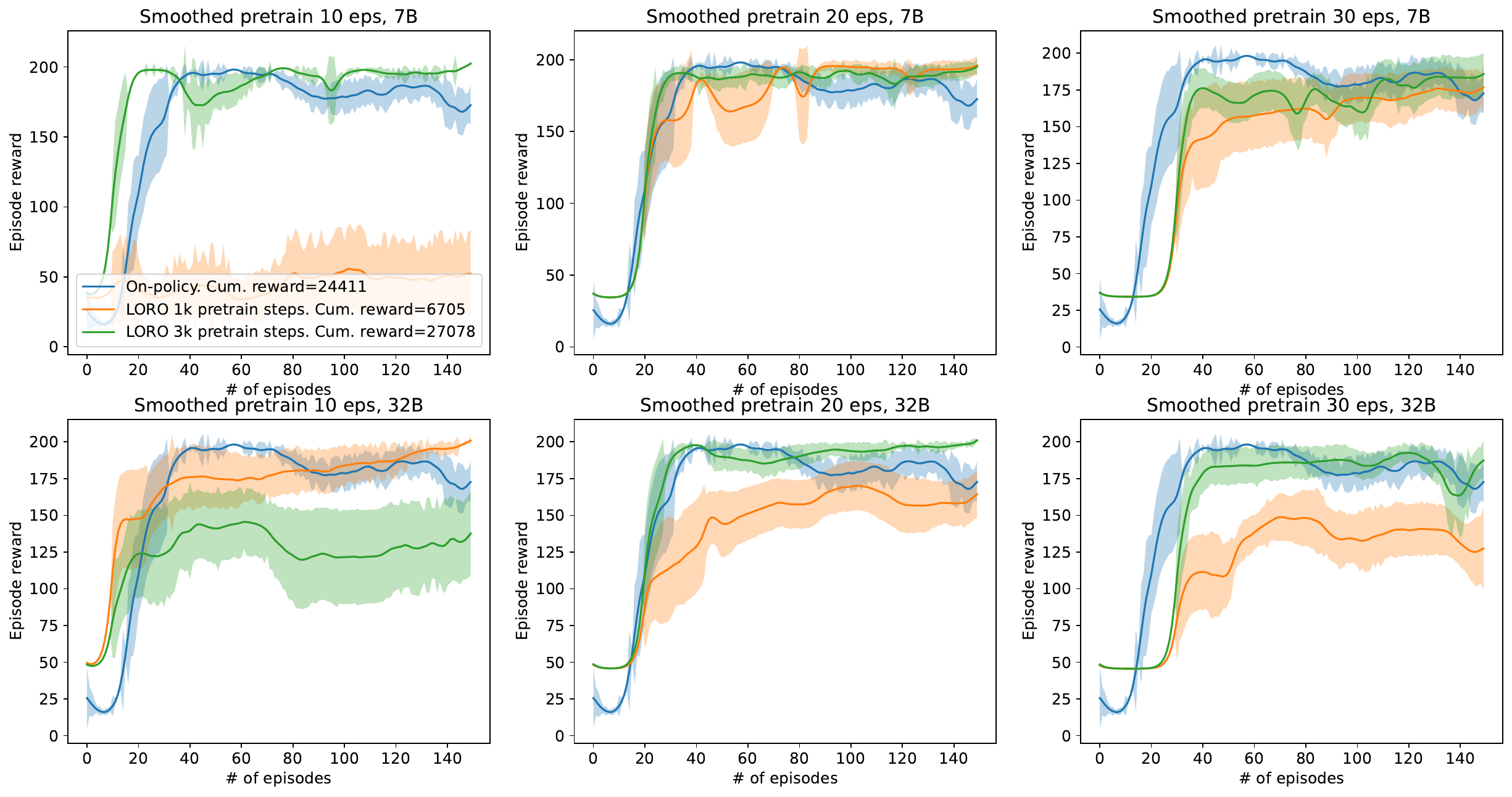}
    \caption{Comparing the effect of different pre-training steps for the CartPole environment.}
\label{fig:cp_pre_steps}
\end{figure}

\begin{figure}[hbtp!]
    \includegraphics[width=\linewidth]{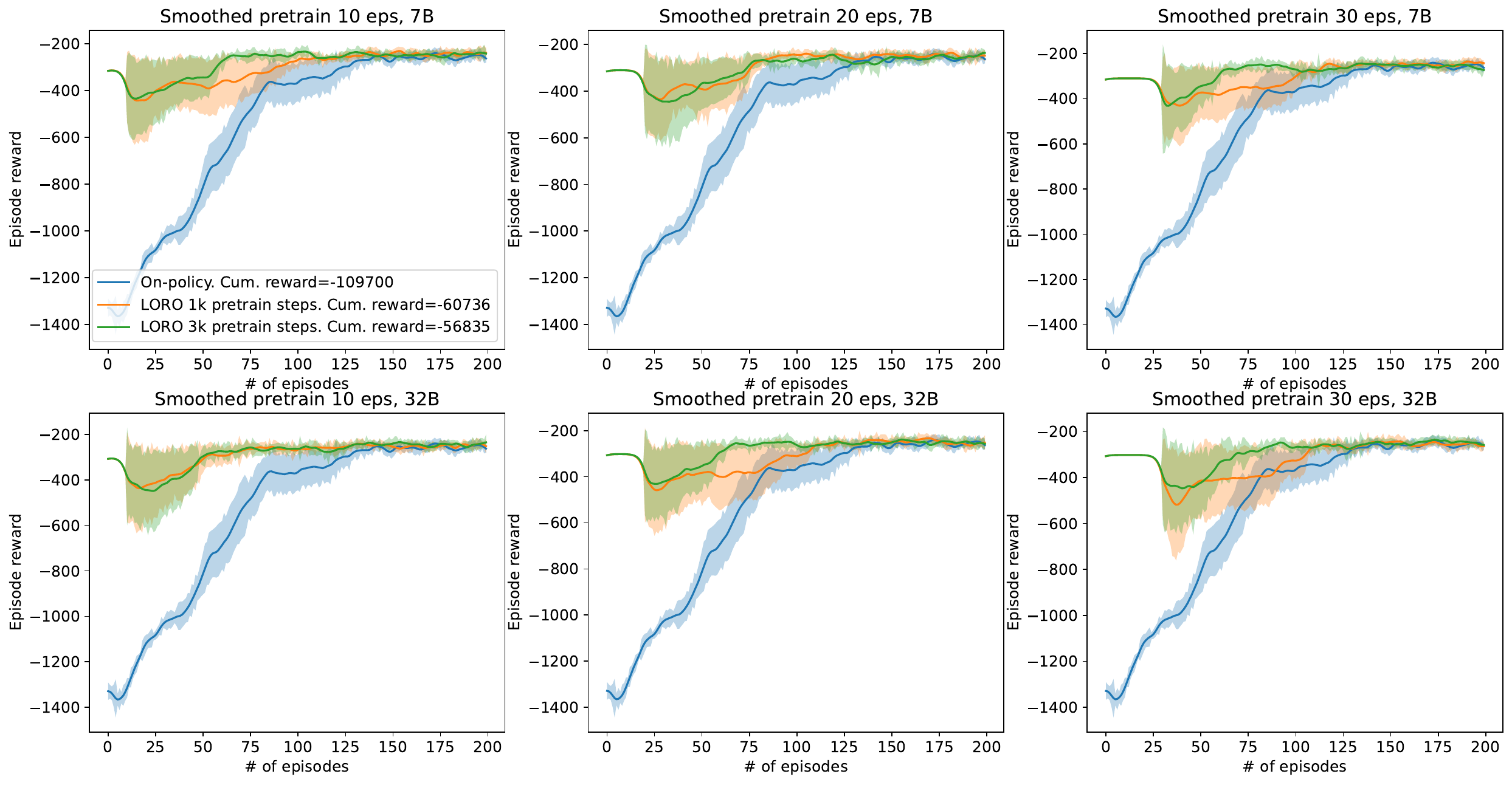}
    \caption{Comparing the effect of different pre-training steps for the Pendulum environment.}
\label{fig:pe_pre_steps}
\end{figure}

\begin{figure}[hbtp!]
    \includegraphics[width=\linewidth]{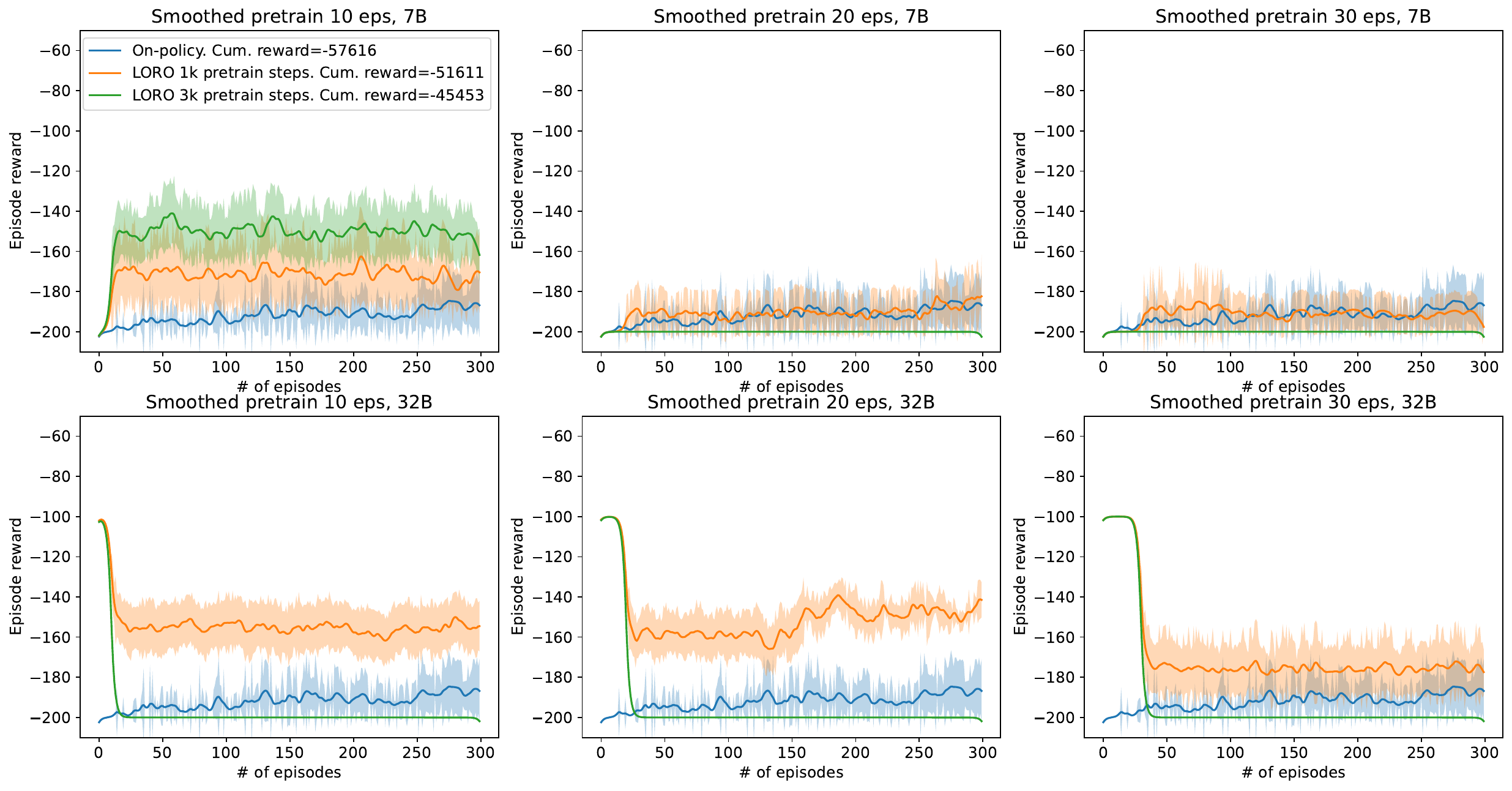}
    \caption{Comparing the effect of different pre-training steps for the MountainCar environment.}
\label{fig:mc_pre_steps}
\end{figure}

\begin{figure}[hbtp!]
    \includegraphics[width=\linewidth]{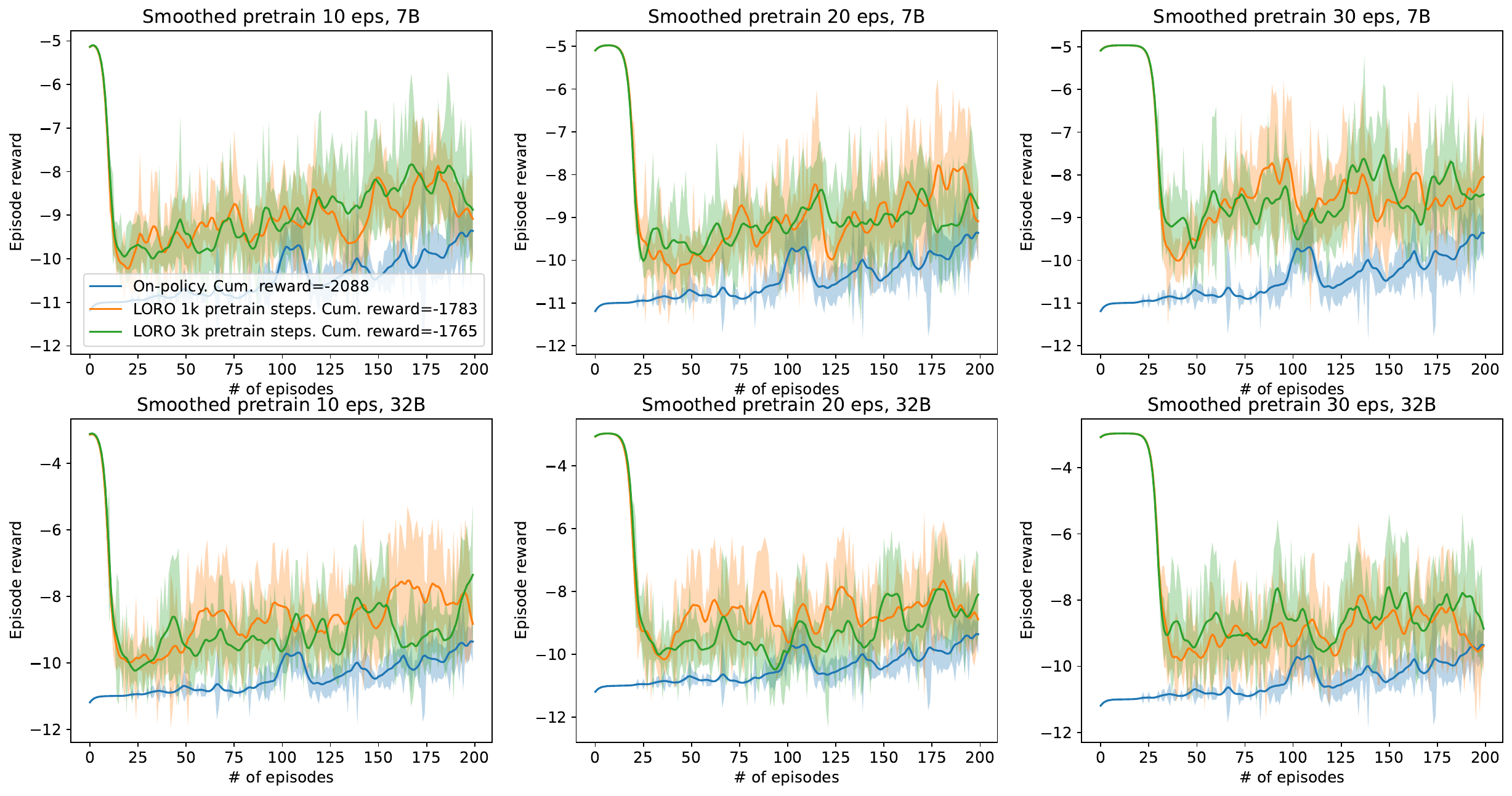}
    \caption{Comparing the effect of different pre-training steps for the Pong environment.}
\label{fig:rp_pre_steps}
\end{figure}

In this section, we evaluate the effect of the number of pre-training steps on the cumulative reward of the policy. We evaluate this with different model sizes and pre-training data on six OpenAI Gym environments and show the result in Figure \ref{fig:cw_pre_steps}, \ref{fig:fl_pre_steps}, \ref{fig:cp_pre_steps}, \ref{fig:pe_pre_steps}, \ref{fig:mc_pre_steps}, \ref{fig:rp_pre_steps}. Overall, we observe no clear advantage of using a higher or lower number of pre-training steps to improve the decision-making quality of the LORO policy.




\newpage
\subsection{Effects of the amount of LLM data}
\label{appendix:pre_eps}

\begin{figure}[hbtp!]
    \includegraphics[width=\textwidth]{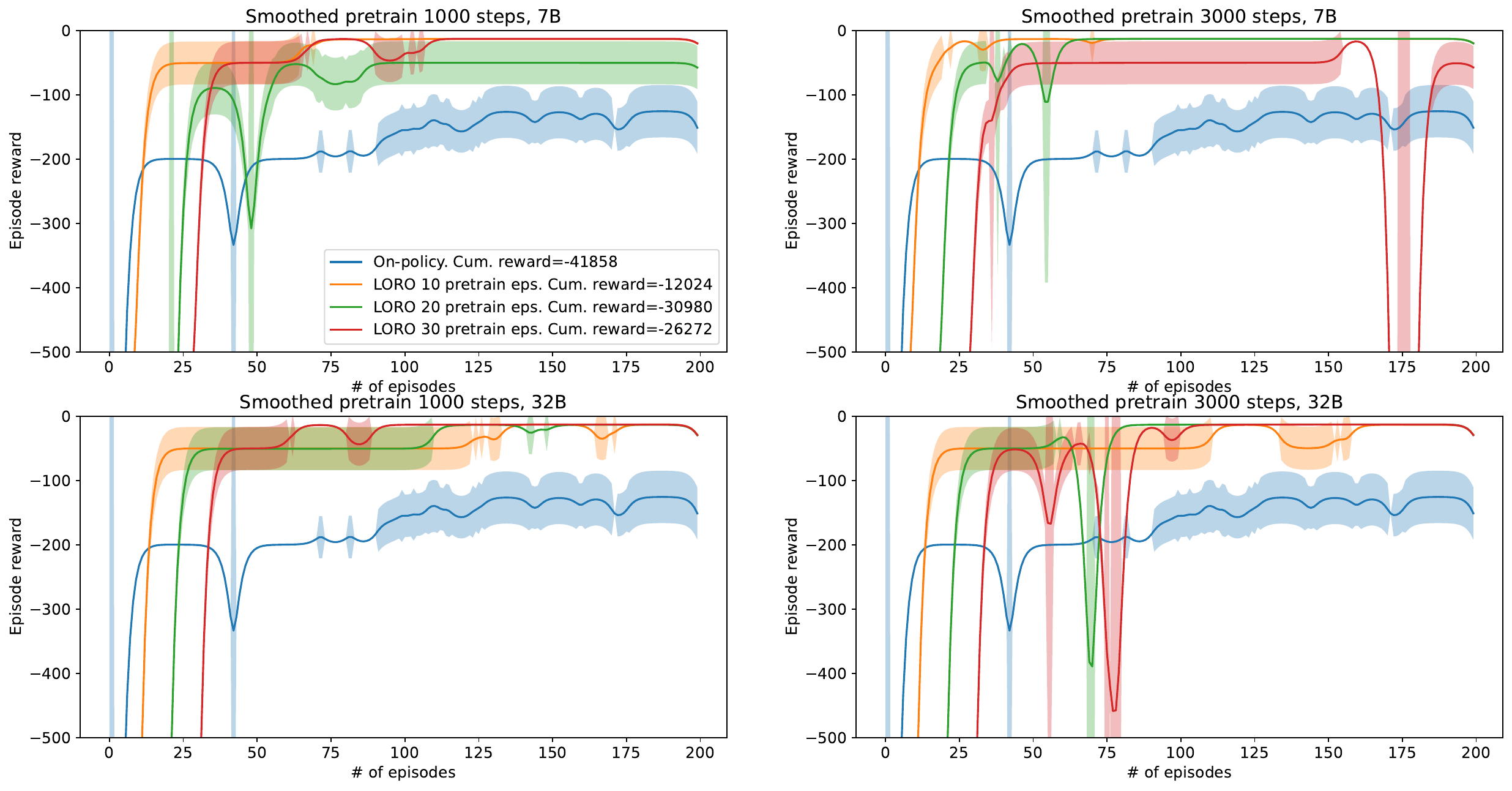}
    \caption{Comparing the effect of different amounts of pre-training data for the CliffWalking environment.}
    \label{fig:cw_pre_eps}
\end{figure}

\begin{figure}[hbtp!]
    \includegraphics[width=\linewidth]{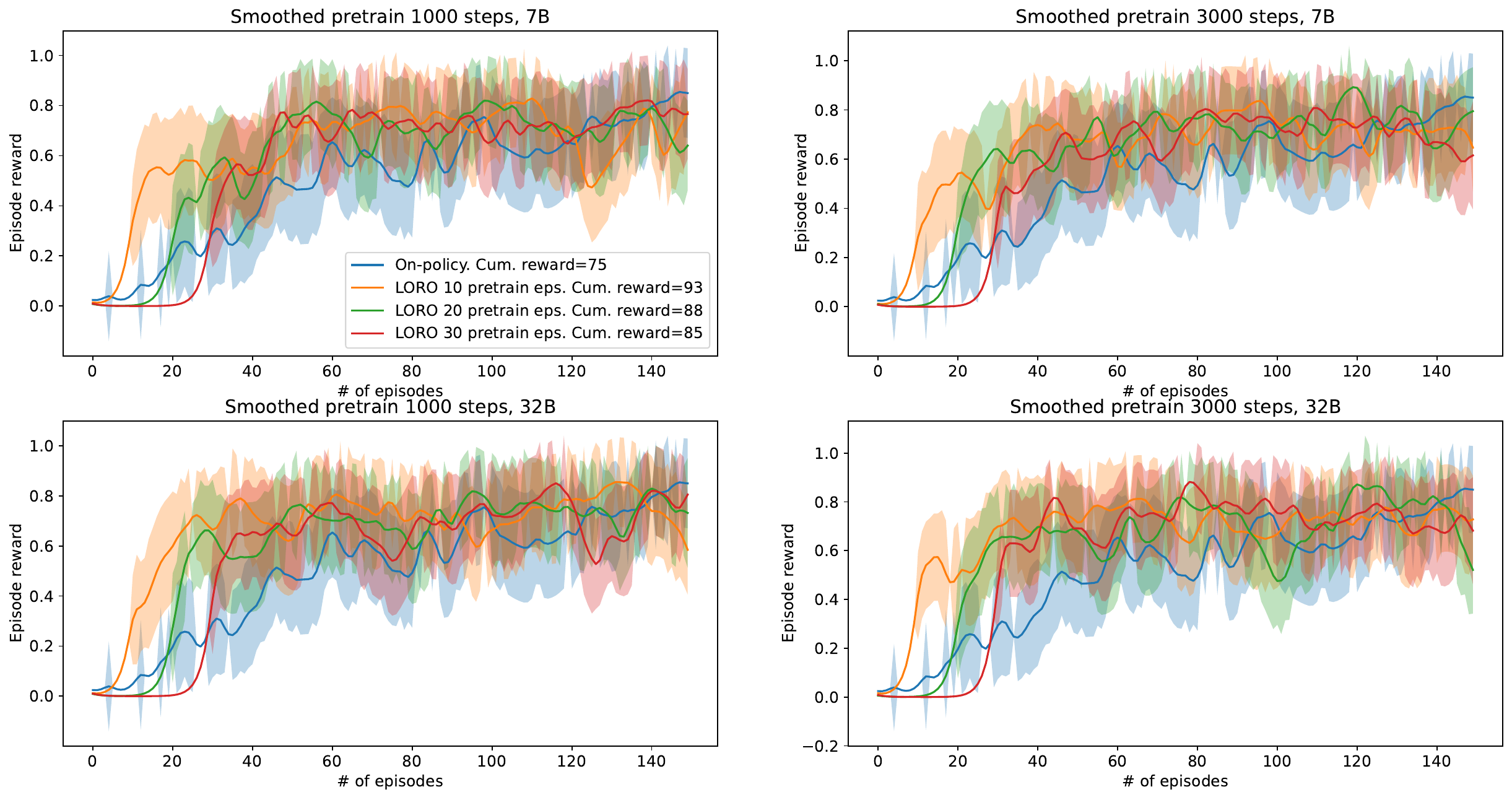}
    \caption{Comparing the effect of different amounts of pre-training data for the FrozenLake environment.}
    \label{fig:fl_pre_eps}
\end{figure}

\begin{figure}[hbtp!]
    \includegraphics[width=\textwidth]{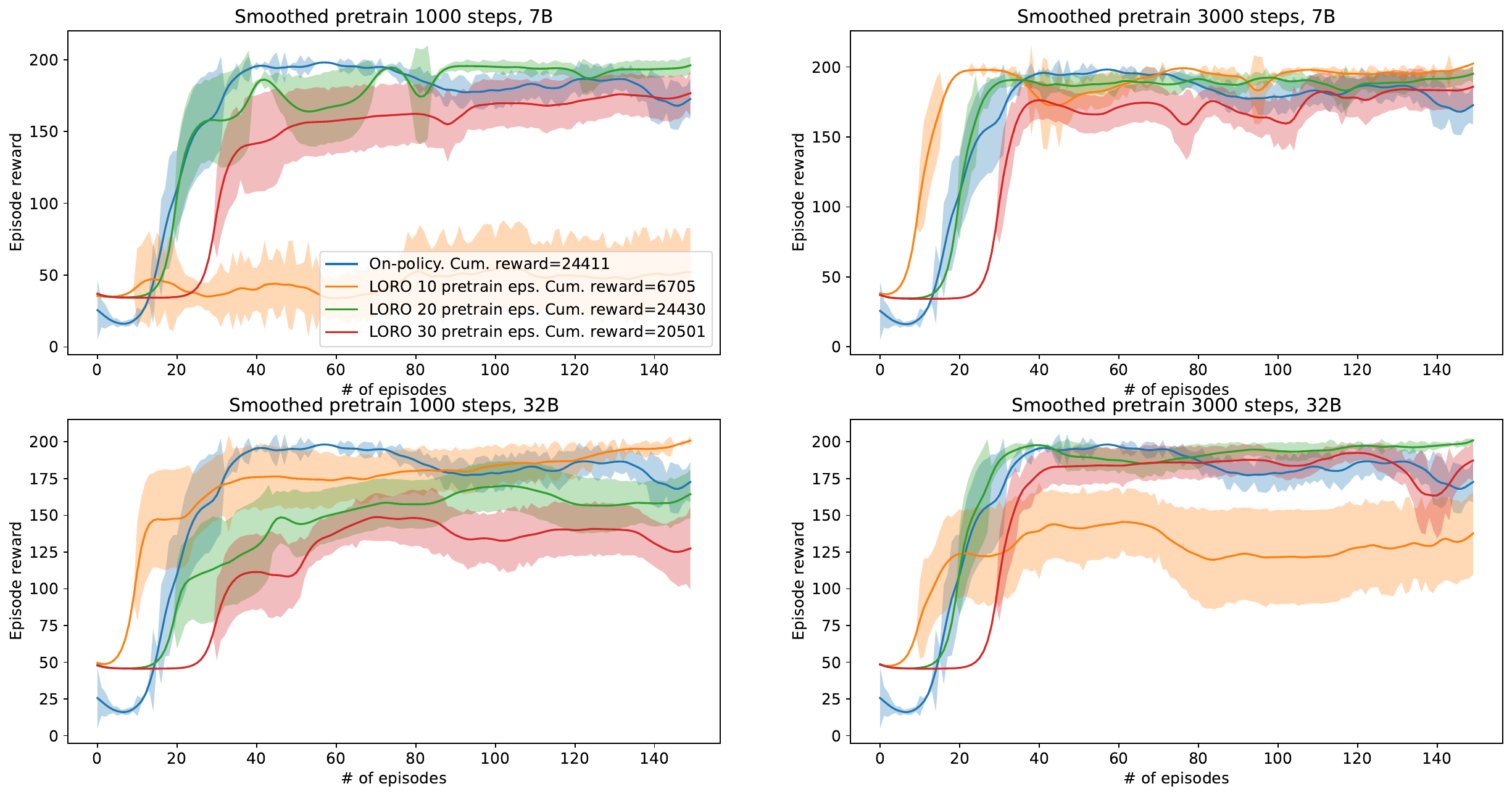}
    \caption{Comparing the effect of different amounts of pre-training data for the CartPole environment.}
    \label{fig:cp_pre_eps}
\end{figure}

\begin{figure}[hbtp!]
    \includegraphics[width=\linewidth]{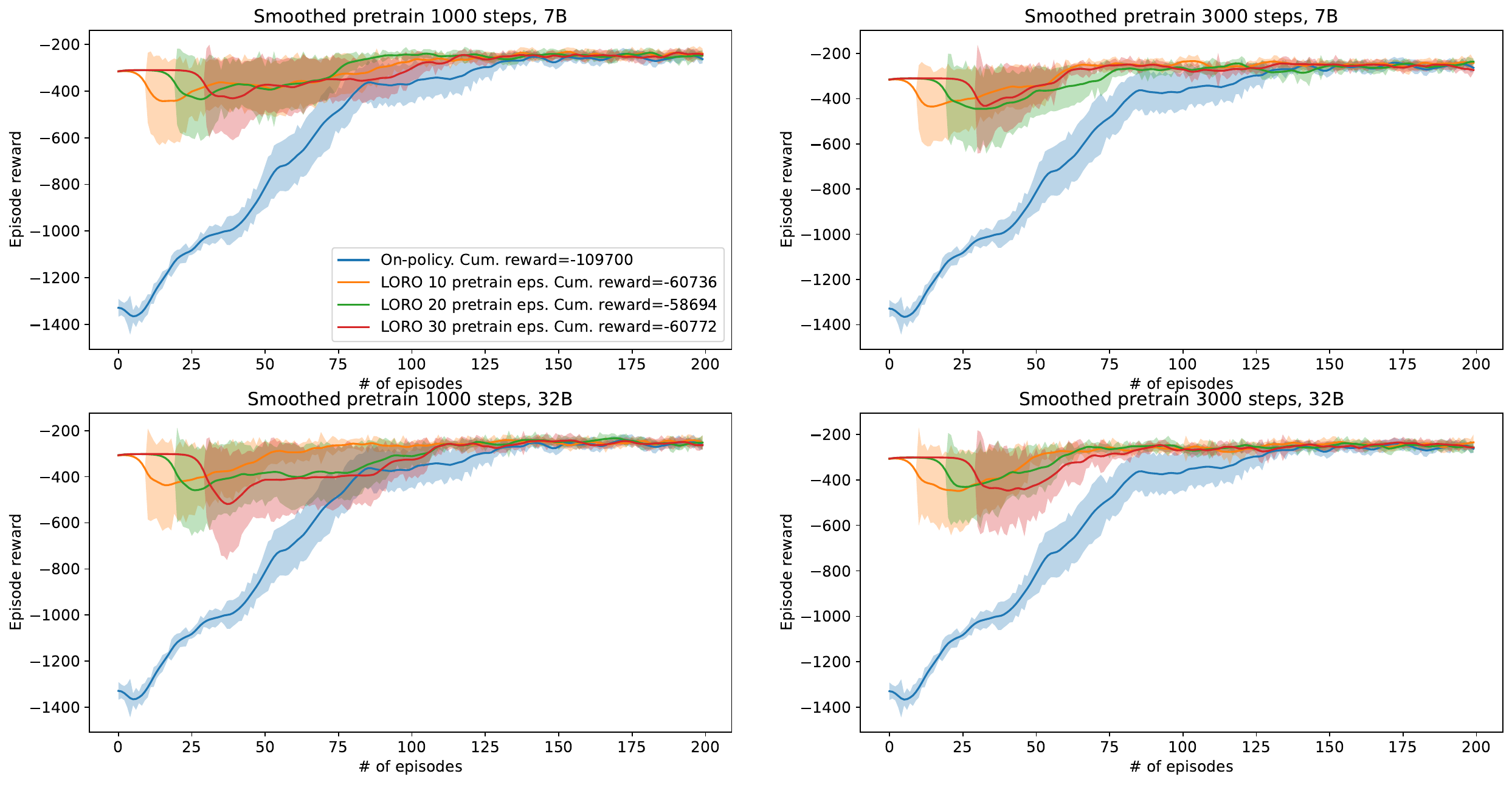}
    \caption{Comparing the effect of different amounts of pre-training data for the Pendulum environment.}
    \label{fig:pe_pre_eps}
\end{figure}

\begin{figure}[hbtp!]
    \includegraphics[width=\linewidth]{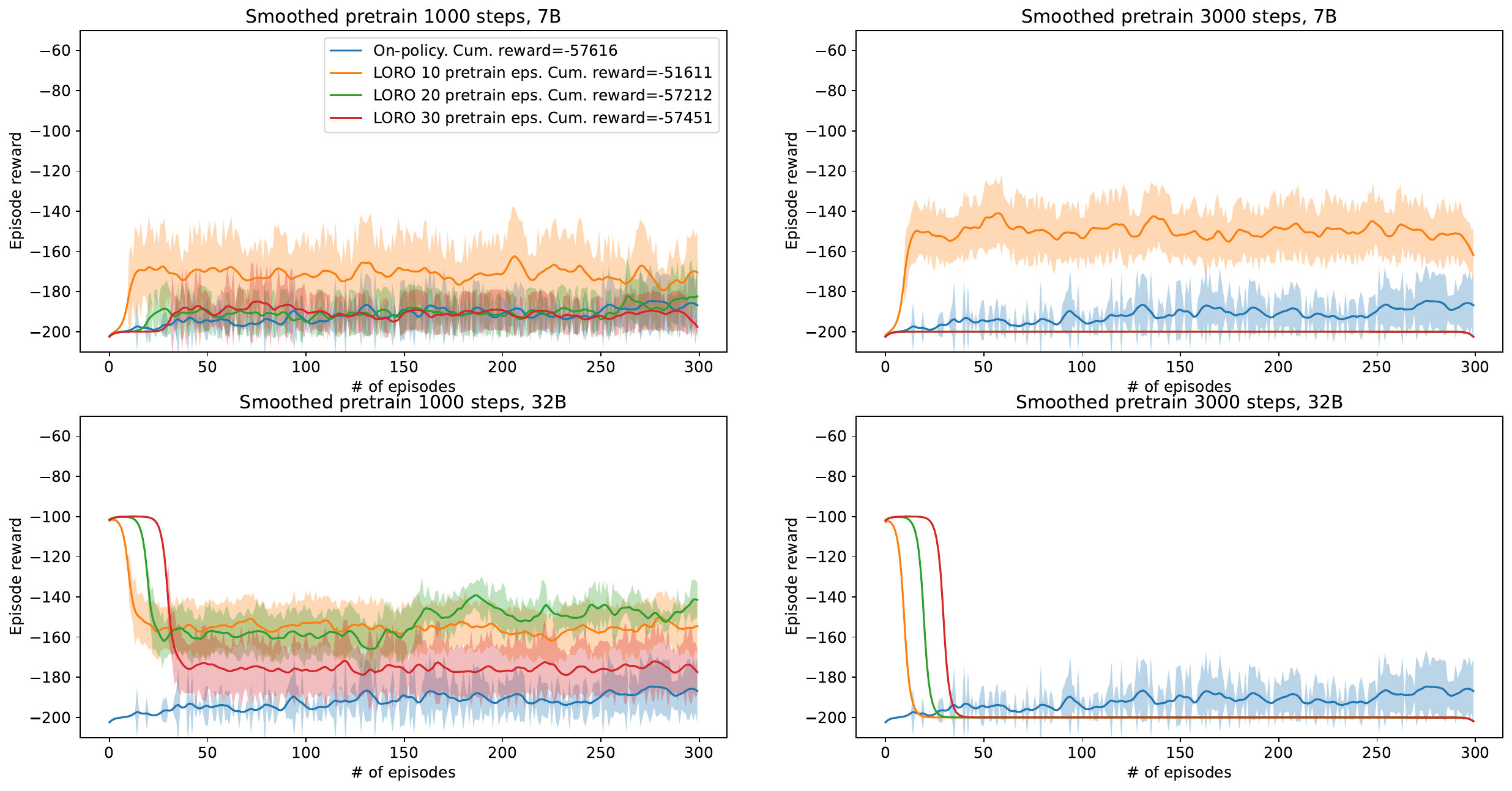}
    \caption{Comparing the effect of different amounts of pre-training data for the MountainCar environment.}
    \label{fig:mc_pre_eps}
\end{figure}

\begin{figure}[hbtp!]
    \includegraphics[width=\linewidth]{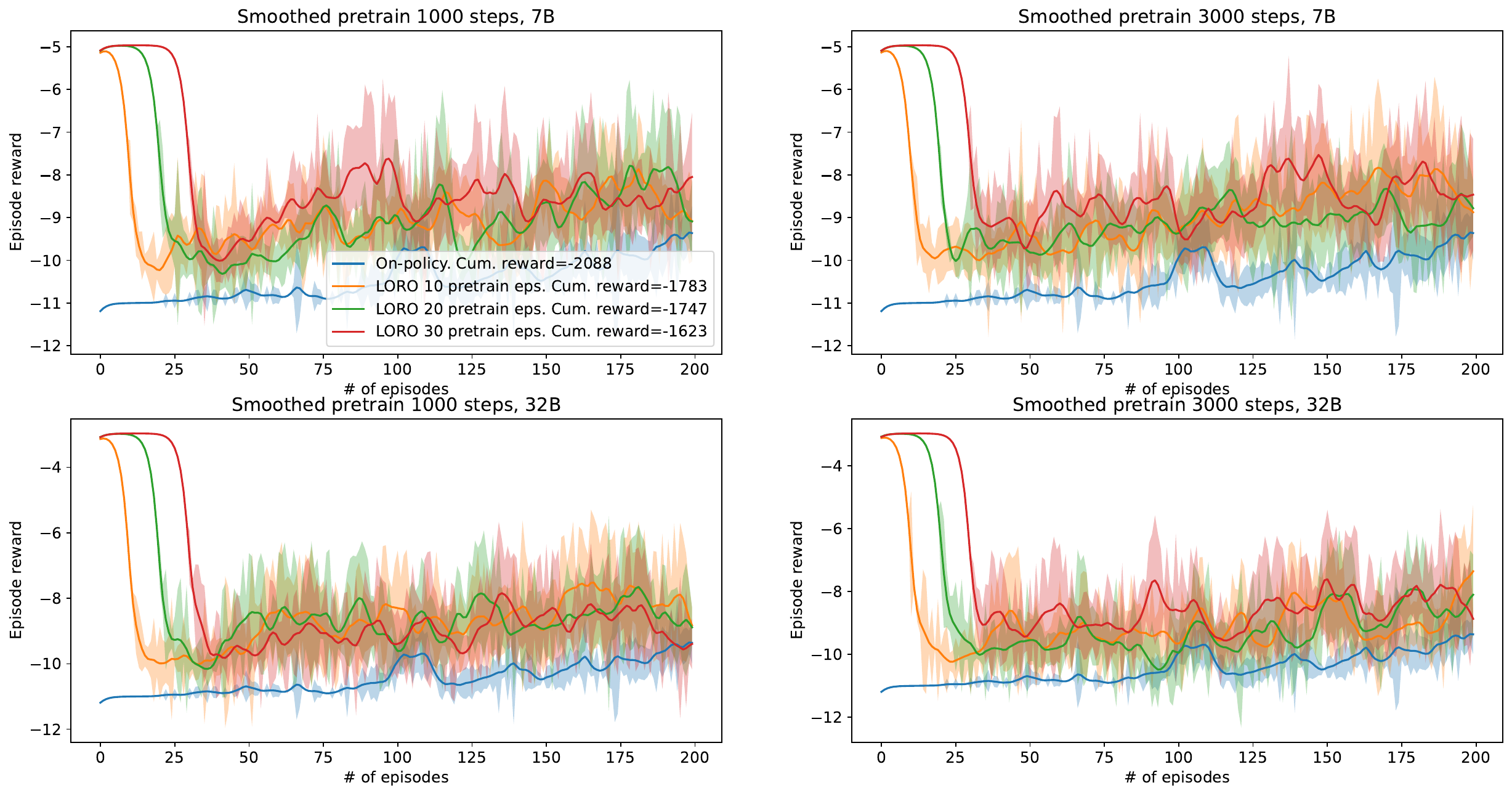}
    \caption{Comparing the effect of different amounts of pre-training data for the Pong environment.}
    \label{fig:rp_pre_eps}
\end{figure}

In this section, we evaluate the effect of the number of pre-training data on the cumulative reward of the policy. We evaluate this with different model sizes and pre-training steps on six OpenAI Gym environments and show the result in Figure \ref{fig:cw_pre_eps}, \ref{fig:fl_pre_eps}, \ref{fig:cp_pre_eps}, \ref{fig:pe_pre_eps}, \ref{fig:mc_pre_eps}, \ref{fig:rp_pre_eps}. Although there exist some differences in the cumulative reward, all baselines converge to a policy with similar performance in a relatively short amount of time. Hence, we observe no clear advantage of using a higher or lower amount of pre-training data to improve the decision-making quality of the LORO policy.




\newpage
\subsection{Effects of SFT and Long CoT}
\label{appendix_sft_lcot}

\begin{figure}[hbtp!]
    \includegraphics[width=\linewidth]{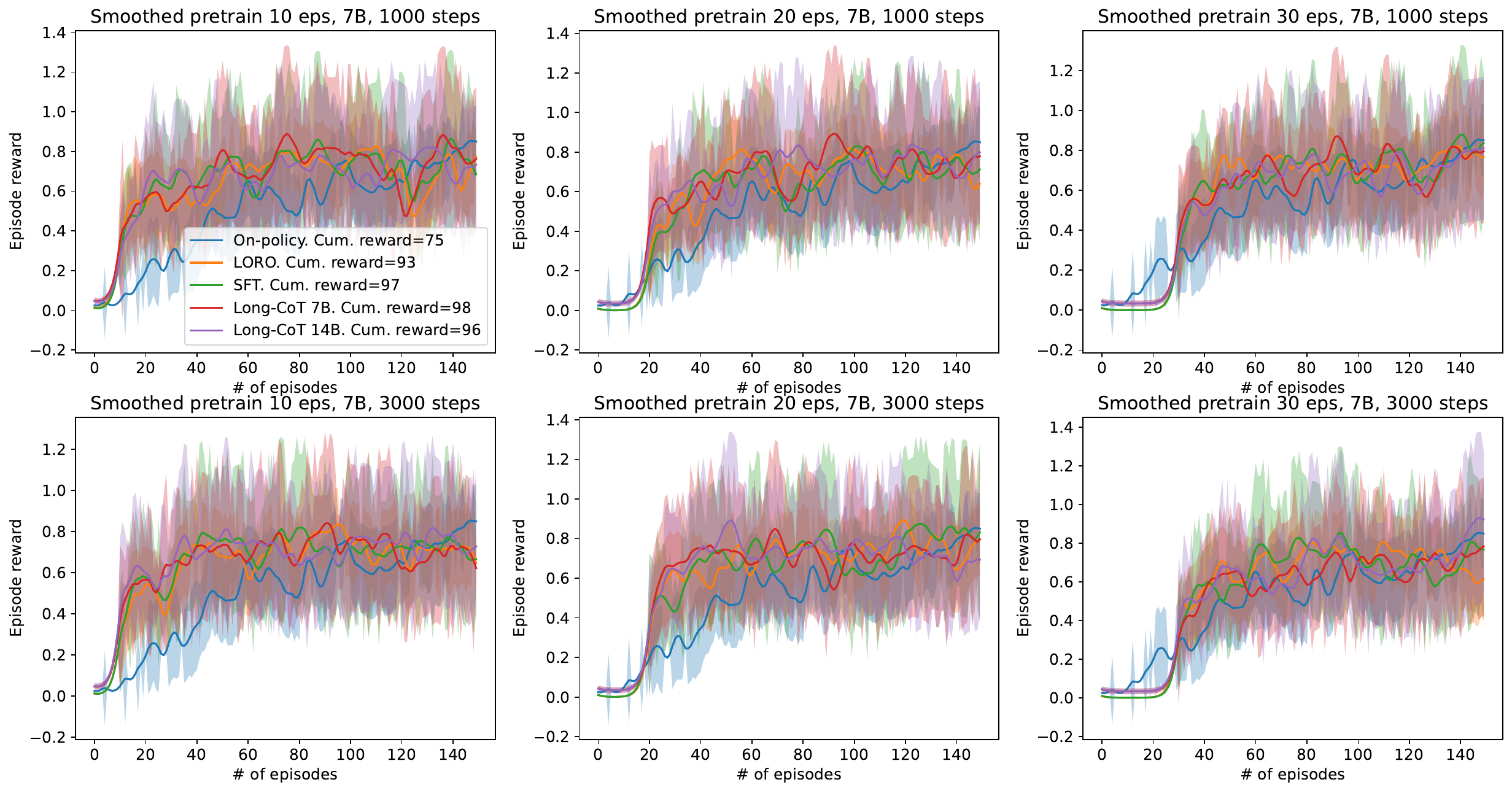}
    \caption{Comparing the effect of Long Chain-of-Thought and Supervised-Fine-Tuning for the FrozenLake environment.}
    \label{fig:fl_sft_lcot}
\end{figure}

\begin{figure}[hbtp!]
    \includegraphics[width=\textwidth]{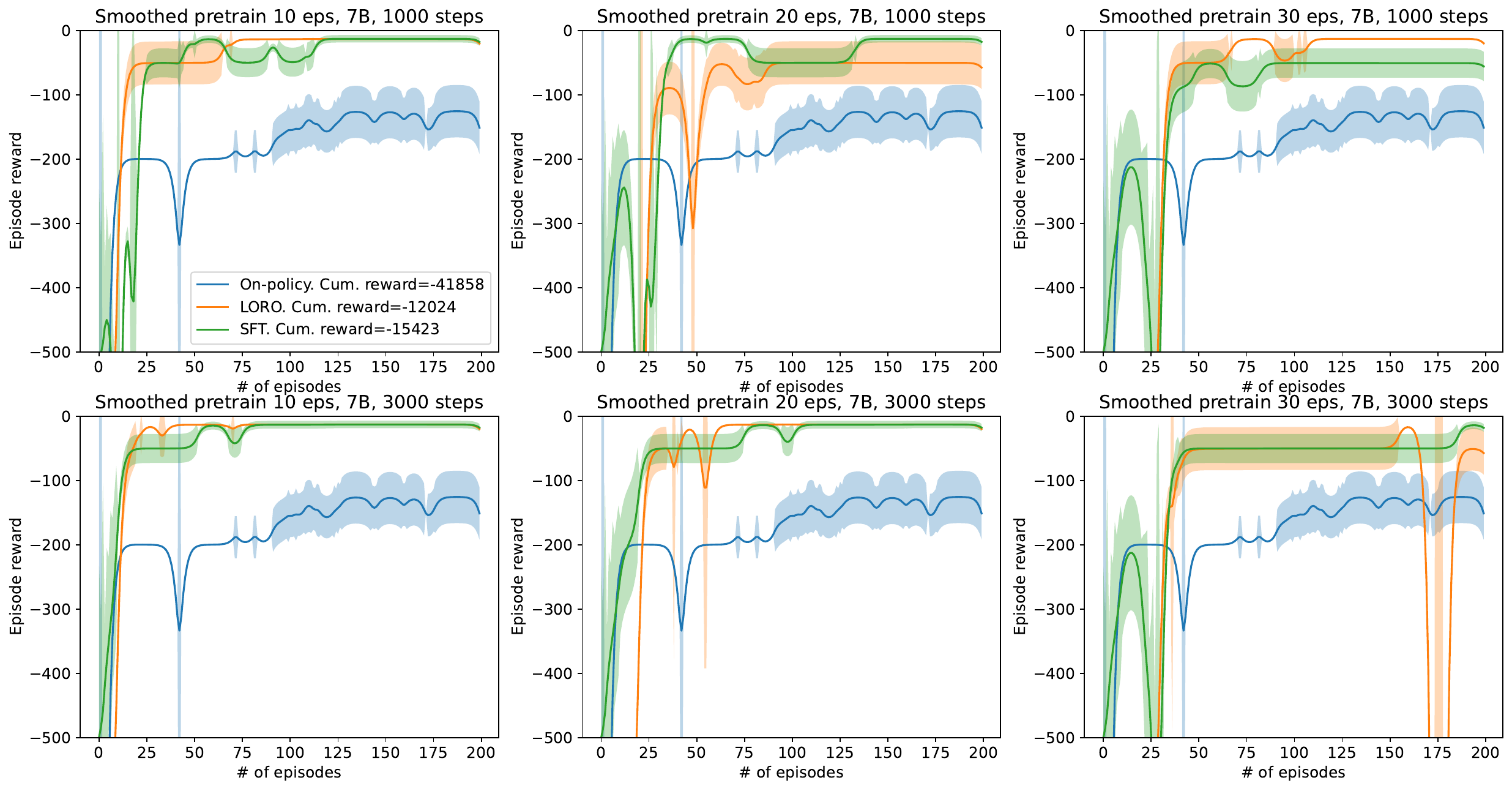}
    \caption{The effect of Supervised-Fine-Tuning for the CliffWalking environment.}
    \label{fig:cw_sft_lcot}
\end{figure}

\begin{figure}[hbtp!]
    \includegraphics[width=\linewidth]{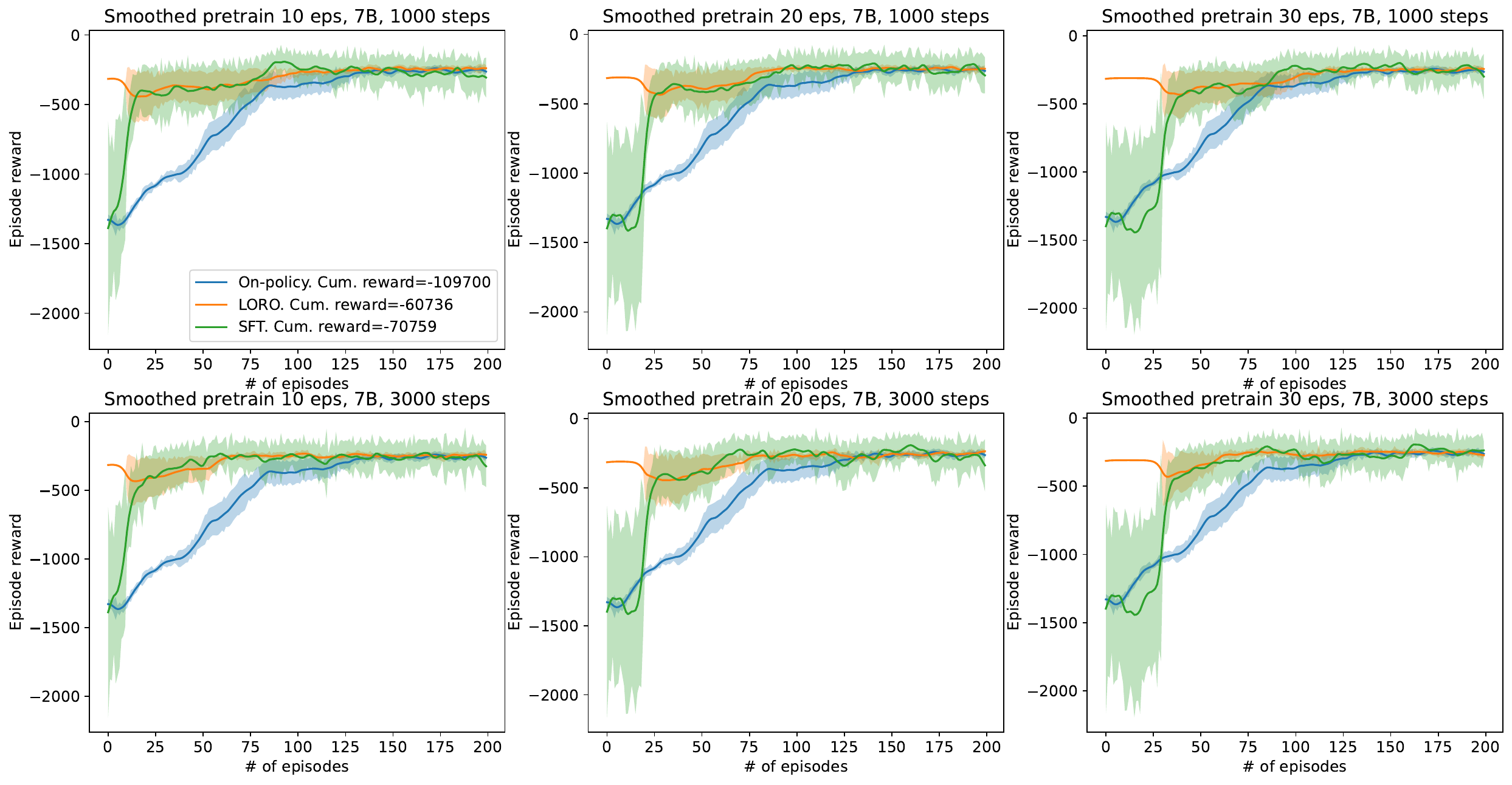}
    \caption{The effect of Supervised Fine-Tuning for the Pendulum environment.}
    \label{fig:pe_sft_lcot}
\end{figure}

In this section, we evaluate the effect of SFT and Long CoT on the cumulative reward of the policy. We evaluate this with different pre-training data and pre-training steps on three OpenAI Gym environments and show the result in Figure \ref{fig:fl_sft_lcot}, \ref{fig:cw_sft_lcot}, \ref{fig:pe_sft_lcot}. Overall, we observe no clear advantage of using SFT and Long CoT over vanilla CoT to improve the decision-making quality of the LORO policy.

\newpage
\subsection{Effects of the history summary}
\label{appendix:env_hist}

For the experiments above, we use an efficient environment history such as ``The holes are in locations: X, Y, Z. You receive zero reward at locations: A, B, C, D''.

\begin{figure}[hbtp!]
    \includegraphics[width=\linewidth]{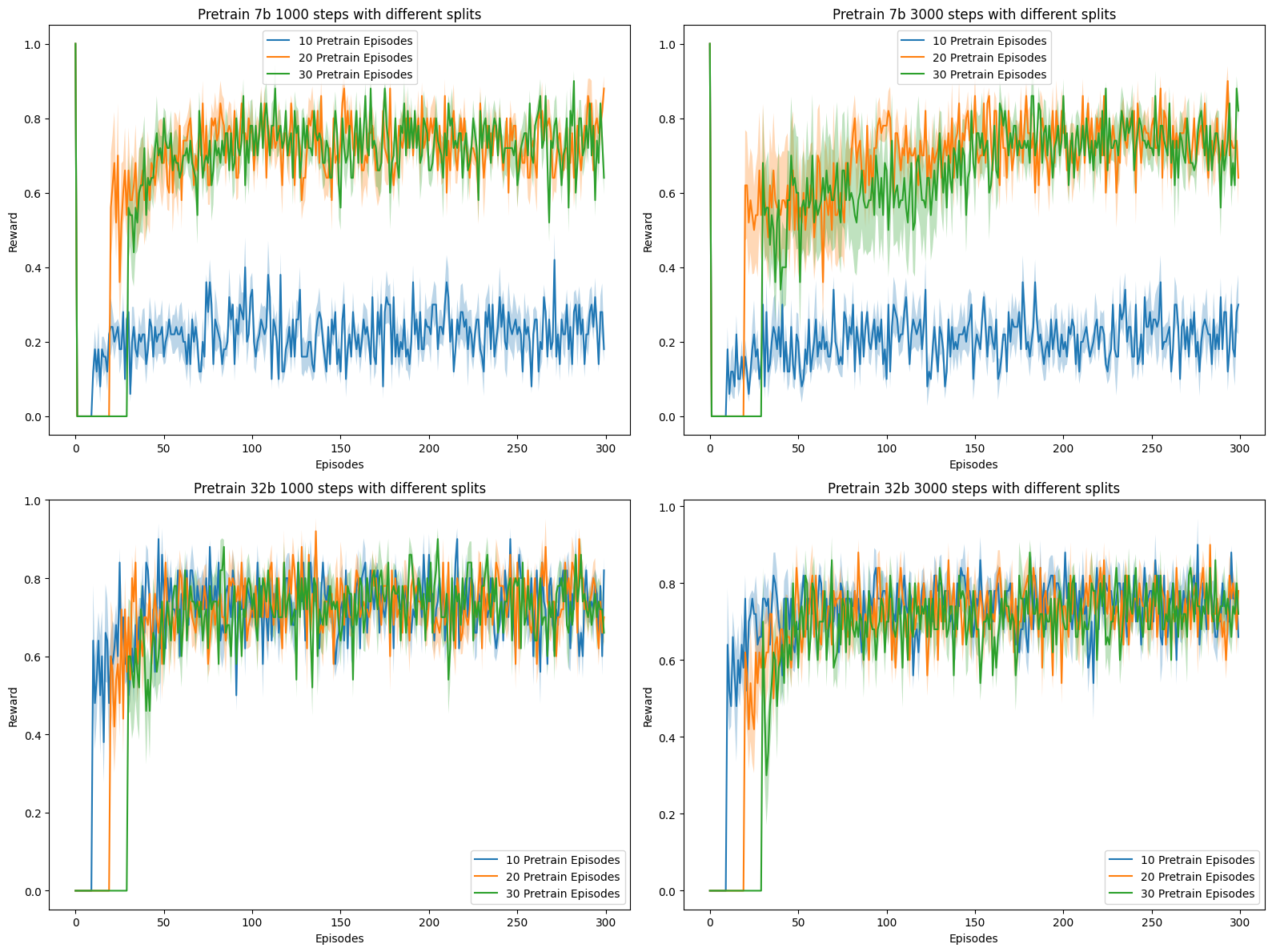}
    \caption{FrozenLake with ineffective environment history.}
    \label{fig:fl_bad_env_hist}
\end{figure}

For the experiment in Figure \ref{fig:fl_bad_env_hist}, we concatenate the observations of each state to the LLM's prompt, with a limited history length: "You visit location X and receive zero reward. You visit location Y and receive one reward. You visit ...".

\newpage
\section{LLM setup}
\label{appendix:llm_setup}

We designed the prompt to choose an action from a list of integers starting from one, since we observed that LLM is more biased toward action zero. After the LLM chooses an action, we extract it by getting the last number returned by the LLM. This design was inherited from  \cite{pandey2024llamagym}, which can be improved since we observe a number of extraction failures from our experiments.

We observe that the vanilla design of LLM, where we ask it only to return the chosen action, performs poorly. Similarly, we implemented and tested the Majority Voting and Best-of-N test-time-scaling methods, but they both perform poorly without CoT.

For all experiments, we limit the generating token to be less than 2000
, top-p 0.6, top-k 0, temperature 0.9. In the SFT experiment, the LLM was trained with LORA \cite{hu2021loralowrankadaptationlarge} with rank 8, alpha 16, dropout 0.05, batch size 1, and using 8-bit quantization.

\section{Prompts}
\label{appendix:prompt}

This section provides a collection of Chain-of-Thought (CoT) prompting examples used across multiple decision-making environments. Each task is presented with a detailed system prompt describing the environment dynamics and action space, followed by a user query that reflects a specific game state. The responses are generated by \texttt{Qwen2.5-7B-Instruct}, illustrating its step-by-step reasoning capabilities. \\

\subsection{Chain-of-Though Prompts}
\subsubsection{Task 1: Cliff Walking}

\promptbox{A Sample of Cliff Walking}{
$\blacktriangleright$ \textbf{System:} \\
You are an expert-level game player. Cliff walking is a task in which you control a player navigating a '4x12' grid world. The ('x', 'y') coordinate indicates the position at row 'x' and column 'y'. The player  needs to find a goal location while avoiding cliffs . The player can choose from 4 actions: move up, move right, move down, or move left. If the player takes an action at ('x', 'y'), he tries to move to ('a', 'b'). Rules: 
\begin{itemize}
    \item 1. If ('a', 'b') is a cliff, the player incurs a large penalty of -100, and is reset to the starting position.
    \item 2. If ('a', 'b') is safe or towards the grid boundary, results in a small penalty of -1. If ('a', 'b') is outside the grid's boundaries, it does not change position but still receive the -1 penalty.
    \item 3. The game ends when the ('a', 'b') is the goal or 200 actions are performed. The goal is to navigate from the starting point to an target , while avoiding the cliff, in as few steps as possible.
\end{itemize}

Type a number to indicate the action. For current position ('x', 'y'), the action means the player try to step into the next position. Type '1' to move up, which means trying to step into ('x-1', 'y'), '2' to move right, which means ('x', 'y+1'), '3' to move down, which means ('x+1', 'y'), or '4' to move left, which means ('x', 'y-1'). Ensure you only provide the action number from the valid action list, i.e., [1, 2, 3, 4]. Environment history: Cliff: Reward -100 at locations: (3, 0). Reward -1 at locations: (2, 0), (3, 0), (1, 0). Previous location: (1, 0), previous action: 1, previous reward: -1.  Return the action at the end of your answer without the target's location.

\vspace{1em}
$\blacktriangleright$ \textbf{User:} \\
You are at location (2, 0) in the grid world.\\Think step by step.\\

\vspace{1em}
$\blacktriangleright$ \textbf{Qwen2.5-7B-Instruct:\\}
1
}

\subsubsection{Task 2: CartPole}

\promptbox{A Sample of CartPole}{
$\blacktriangleright$ \textbf{System:} \\
You are an expert-level game player. In the CartPole game, you control a cart that moves along a horizontal track. There is a pole standing upright on the cart. The goal of the game is to keep the pole balanced upright by moving the cart left or right. The game ends if the pole tilts too far from the vertical position or if the cart moves too far from the center of the track. The longer you can keep the pole balanced, the higher your score.Note that when the Cart Position is out of the (-2.4, 2.4) zone or the Pole Angle is out of the zone (-.2095, .2095), the round ends and the game is lost. The goal is to keep the pole balanced upright for as long as possible. \\Type '1' to push the cart to the left or '2' to push the cart to the right. Ensure you only provide the action number from the valid action list, i.e., [1, 2].

\vspace{1em}
$\blacktriangleright$ \textbf{User:} \\
The cart is positioned at 0.006, with a velocity of 0.04 towards the right. The pole is tilted at 0.02 radians, rotating at 0.02 radians per second towards the right.\\
Think step by step.

\vspace{1em}
$\blacktriangleright$ \textbf{Qwen2.5-7B-Instruct:} \\
Given the current state of the CartPole game:
\begin{itemize}
  \item Cart Position: 0.006 (slightly to the right of the center)
  \item Cart Velocity: 0.04 (moving slightly to the right)
  \item Pole Angle: 0.02 radians (slightly tilted to the right)
  \item Pole Angular Velocity: 0.02 radians/second (tilting to the right)
\end{itemize}

Let's analyze the situation step by step:
\begin{enumerate}
  \item \textbf{Cart Position}: The cart is very close to the center, at 0.006. This is within the safe range, so we don't need to move it significantly.
  \item \textbf{Cart Velocity}: The cart is moving slightly to the right, which is good because it helps counteract the tilt of the pole.
  \item \textbf{Pole Angle}: The pole is tilted 0.02 radians to the right. This is a small tilt, but it indicates that the pole is starting to lean. We need to act quickly to prevent it from falling over.
  \item \textbf{Pole Angular Velocity}: The pole is rotating at 0.02 radians/second to the right, which means it is accelerating its tilt. This suggests that the cart's movement is not sufficient to stabilize the pole.
\end{enumerate}

Given these observations, we should try to slow down the pole's tilt by moving the cart to the left. This will help counteract the rightward tilt of the pole.\\

\textbf{Action: 1 (Push the cart to the left)}\\

This action will help to stabilize the pole by moving the center of mass back towards the center, which should reduce the angular velocity of the pole.
}

\subsubsection{Task 3: Mountain Car}

\promptbox{A Sample of Mauntain Car}{
$\blacktriangleright$ \textbf{System:} \\
You are an expert-level game player. In the Mountain Car game, you control a car placed stochastically at the bottom of a sinusoidal valley. The only possible actions are the accelerations that can be applied to the car in either direction. The goal of the game is to strategically accelerate the car to reach the goal state on top of the right hill as quickly as possible. The episode ends if either the car reaches the goal position on top of the right hill or the length of the episode is 200. The goal is to reach the flag placed on top of the right hill as quickly as possible. \\
Type '1' to accelerate to the left, '2' to not accelerate, or '3' to accelerate to the right.Ensure you only provide the action number from the valid action list, i.e., [1, 2, 3].

\vspace{1em}
$\blacktriangleright$ \textbf{User:} \\
The car is positioned at -0.541, with a velocity of 0.000 towards the left.\\
Think step by step.

\vspace{1em}
$\blacktriangleright$ \textbf{Qwen2.5-7B-Instruct:} \\
Given the car's position at -0.541 and a velocity of 0.000 towards the left, the car is slightly to the left of the center of the valley. Since the car is not moving yet and is close to the middle, accelerating to the right (action 3) might help gain some speed towards the right hill without overshooting too much.\\

Action: 3
}

\subsubsection{Task 4: FrozenLake}

\promptbox{A Sample of FrozenLake (CoT)}{
$\blacktriangleright$ \textbf{System:} \\
You are an expert-level game player. In the FrozenLake game, the player starts at the start position of the grid and tries to reach the goal position . There are holes which the player must avoid. The frozen lake is slippery, meaning that the player might not always move in the intended direction. The game ends when the player reaches the goal or falls into a hole. The goal is to navigate across the frozen lake and reach the goal position without falling into any holes. For current position ('x', 'y'), the action means the player try to step into the next position. The possible actions are: 
\begin{itemize}
  \item 1: Move left, which means ('x', 'y-1'),
  \item 2: Move down, which means ('x+1', 'y'),
  \item 3: Move right, which means ('x', 'y+1'),
  \item 4: Move up, which means trying to step into ('x-1', 'y').
\end{itemize}
Ensure you only provide the action number from the valid action list, i.e., [1, 2, 3, 4]. Do not return the target's coordination. Environment history: Reward 0.0 at locations: (0, 0). Previous location: (0, 0), previous action: 1, previous reward: 0.0.  Return the action at the end of your answer without the target's location. 

\vspace{1em}
$\blacktriangleright$ \textbf{User:} \\
You are at row 0, column 0. \\
Think step by step.

\vspace{1em}
$\blacktriangleright$ \textbf{Qwen2.5-7B-Instruct:} \\
Since the current position is (0, 0) and the previous action was moving left (which is not possible as it would take you out of bounds), the only feasible action to consider is moving down to (0, 1). Therefore, the action is: \textbf{2}.
}

\promptbox{A Sample of FrozenLake (Long CoT)}{
$\blacktriangleright$ \textbf{System:} \\
You are an expert-level game player. In the FrozenLake game, the player starts at the start position of the grid and tries to reach the goal position . There are holes which the player must avoid. The frozen lake is slippery, meaning that the player might not always move in the intended direction. The game ends when the player reaches the goal or falls into a hole. The goal is to navigate across the frozen lake and reach the goal position without falling into any holes. For current position ('x', 'y'), the action means the player try to step into the next position. The possible actions are: 
\begin{itemize}
  \item 1: Move left, which means ('x', 'y-1'),
  \item 2: Move down, which means ('x+1', 'y'),
  \item 3: Move right, which means ('x', 'y+1'),
  \item 4: Move up, which means trying to step into ('x-1', 'y').
\end{itemize}
Ensure you only provide the action number from the valid action list, i.e., [1, 2, 3, 4]. Do not return the target's coordination. Environment history: Reward 0.0 at locations: (0, 0). Previous location: (0, 0), previous action: 1, previous reward: 0.0.  Return the action at the end of your answer without the target's location. 

\vspace{1em}
$\blacktriangleright$ \textbf{User:} \\
The current position of the You are at row 0, column 0. \\
Think step by step. 

\vspace{1em}
$\blacktriangleright$ \textbf{DeepSeek-R1-Distill-Qwen-7B:} \\

Okay, so I'm trying to figure out where to move in this FrozenLake game. I'm at position (0,0), which is the top-left corner of the grid. The goal is to reach the goal position, which I think is at the bottom-right corner, maybe (4,4) since it's a 5x5 grid. But I'm not entirely sure, so I'll assume that's the case.

First, I need to consider the possible actions. The actions are numbered 1 to 4, corresponding to moving left, down, right, and up respectively. Each action has a direction, but because the ice is slippery, I might not end up where I intended. So, I have to plan carefully to avoid falling into holes.\\
\\

Let me think about each action:
\begin{itemize}
    \item Action 1: Move left. But wait, I'm already at column 0, which is the leftmost column. So moving left from here would take me off the grid, which is probably a hole. So I shouldn't choose action 1.
    \item Action 2: Move down. That would take me to (1,0). I need to check if that's a hole. From the grid, I remember that some cells are safe and some are holes. Let me visualize the grid: 
    \begin{itemize}
        \item Row 0: S F F F F
        \item Row 1: F F H F F
        \item Row 2: F H F F F
        \item Row 3: F F F H F
        \item Row 4: F F F F G
    \end{itemize}
    Wait, I'm not sure if that's accurate, but I think row 1, column 2 is a hole (H), and row 2, column 1 is another hole. So moving down to (1,0) is safe because row 1, column 0 is F (safe). So action 2 seems safe.
    \item Action 3: Move right. That would take me to (0,1). I think that's safe too because row 0, column 1 is F.
    \item Action 4: Move up. But I'm at row 0, so moving up would take me off the grid, which is a hole. So action 4 is out.
\end{itemize}

So, from (0,0), the valid actions are 2 and 3. I need to choose between them. Let's think about the long-term implications.

If I go down to (1,0), then from there, possible actions are:
\begin{itemize}
    \item Move left: off grid (hole)
    \item Move down: (2,0) which is safe
    \item Move right: (1,1) which is safe
    \item Move up: back to (0,0)
\end{itemize}

If I go right to (0,1), then from there:
\begin{itemize}
    \item Move left: back to (0,0)
    \item Move down: (1,1) which is safe
    \item Move right: (0,2) which is safe
    \item Move up: off grid (hole)\\
\end{itemize}

Hmm, both paths seem viable. Maybe I should go right to (0,1) because moving down from there might lead me closer to the goal without getting stuck. Alternatively, going down might allow me to move right or left without immediate issues.\\
\\

Wait, but if I go down to (1,0), then moving right to (1,1) is safe, and from there, moving down again to (2,1) is a hole because row 2, column 1 is H. Oh no, that's a problem. So from (1,0), if I move right to (1,1), then down to (2,1) is a hole. So that's bad.

Alternatively, from (1,0), if I move down to (2,0), which is safe, then from there, moving right to (2,1) is a hole. So again, that's a problem.\\
\\

Wait, maybe I'm misremembering the grid. Let me try to recall the layout. I think row 1, column 2 is H, and row 2, column 1 is H. So, from (1,0), moving right goes to (1,1), which is safe, but then moving down from there would go to (2,1), which is H. So that's a problem.

Alternatively, from (1,0), moving down to (2,0) is safe, but then moving right to (2,1) is H. So either way, moving down first might lead to a hole in the next step.

So maybe going right to (0,1) is better. From (0,1), moving down to (1,1) is safe, and then from there, moving right to (1,2) is H, which is bad. Or moving right to (0,2), which is safe, then from there, moving down to (1,2) is H. So again, same issue.\\
\\

Wait, maybe I'm overcomplicating this. Perhaps I should try to find a path that avoids the holes. Let's think about possible paths.

From (0,0), go right to (0,1). From there, go down to (1,1). From (1,1), go right to (1,2), but that's H. So that's bad. Alternatively, from (1,1), go right to (1,2) is H, so maybe go down to (2,1), which is H. So that's no good.

Alternatively, from (0,1), go down to (1,1), then from there, go right to (1,2) is H, or left to (1,0), which is back where I came from. Hmm.\\
\\

Wait, maybe from (0,1), go right to (0,2). From (0,2), go down to (1,2) is H. So that's bad.

Alternatively, from (0,1), go down to (1,1), then from there, go right to (1,2) is H, or left to (1,0), which is safe. So from (1,1), moving left to (1,0) is safe, but then from (1,0), moving down to (2,0) is safe, and from there, moving right to (2,1) is H. So that's a problem.

This is getting complicated. Maybe I should try a different approach. Let's think about the possible paths step by step.

From (0,0), options are 2 (down) and 3 (right).

If I go down to (1,0):
\begin{itemize}
    \item From (1,0), possible actions: down to (2,0), right to (1,1)
    \item If I go down to (2,0):
    \begin{itemize}
        \item From (2,0), possible actions: right to (2,1) is H, so that's bad. So maybe go right to (2,1) is H, so can't go there. So from (2,0), the only safe move is right to (2,1) is H, so maybe move up to (1,0), but that's where I came from. Or move right to (2,1) is H, so that's not allowed. So from (2,0), the only safe move is to go right to (2,1), but that's H, so that's bad. Alternatively, move left to (2,-1), which is off grid, so that's a hole. So from (2,0), I can't move anywhere without falling into a hole. So that's a dead end.
    \end{itemize}
\end{itemize}

So going down to (1,0) leads to a dead end because from (2,0), I can't move anywhere safely.\\

So, maybe I should go right to (0,1) instead.

From (0,1):
\begin{itemize}
    \item Possible actions: down to (1,1), right to (0,2)
    \item If I go down to (1,1):
    \begin{itemize}
        \item From (1,1), possible actions: down to (2,1) is H, right to (1,2) is H, left to (1,0), up to (0,1)
        \item So, from (1,1), the only safe move is left to (1,0) or up to (0,1). If I go left to (1,0), then from there, I can go down to (2,0), but from (2,0), as before, it's a dead end. Alternatively, from (1,1), go up to (0,1), which is where I came from, so that's not helpful.
    \end{itemize}
    \item If I go right to (0,2):
    \begin{itemize}
        \item From (0,2), possible actions: down to (1,2) is H, right to (0,3), left to (0,1)
        \item So, from (0,2), the safe moves are right to (0,3) or left to (0,1). If I go right to (0,3), then from there, I can go down to (1,3), which is safe, and so on.
    \end{itemize}
\end{itemize}

So, from (0,1), going right to (0,2) seems better because it allows me to move further right without hitting a hole immediately.\\
\\

So, let's try that path:

\textbf{Action: 1}
}

\subsubsection{Task 5: Pong}

\promptbox{A Sample of Pong}{
$\blacktriangleright$ \textbf{System:} \\
You are an expert-level game player. In the Pong game, you play the ball with your opponent, each player rallys the ball by moving the paddles on the playfield. Paddles move only vertically on the playfield. A player scores one point when the opponent hits the ball out of bounds or misses a hit. The first player to score 21 points wins the game. The number of frameskip is set to 4.  The goal is to knock out your opponent. Type 1 for NOOP (no operation), 3 to move up, 4 to move down. Ensure you only provide the action number from the valid action list, i.e., [1, 3, 4].

\vspace{1em}
$\blacktriangleright$ \textbf{User:} \\
The origin (0,0) is in the top left corner. You are at position (188, 78), your opponent is at position (64, 8), the ball is at (0, 0), the ball velocity is (0, -60). Your opponent's score is 0, your score is 0.\\
Think step by step.

\vspace{1em}
$\blacktriangleright$ \textbf{Qwen2.5-7B-Instruct:} \\
Given the current state of the game:
\begin{itemize}
    \item Your paddle is at position (188, 78).
    \item Your opponent's paddle is at position (64, 8).
    \item The ball is at position (0, 0) and is moving upwards with a velocity of (0, -60).
\end{itemize}

\textbf{Step-by-step reasoning:}
\begin{itemize}
    \item The ball is at the top of the playfield and is moving downwards. It will be near the middle of the playfield soon.
    \item With a frameskip of 4, the opponent will not react until the ball has moved significantly downward.
    \item Since the ball is about to come towards you, you should prepare to move your paddle to intercept the ball.
\end{itemize}

\textbf{Action Plan:}
\begin{itemize}
    \item Move your paddle up to intercept the ball as it comes towards you. This ensures that you can return the ball effectively.
\end{itemize}

\textbf{Action: 3} (move up)

By moving up, you position yourself to intercept the ball as it comes towards the center of the playfield.
}

\subsubsection{Task 6: Pendulum}

\promptbox{A Sample of Pendulum}{
$\blacktriangleright$ \textbf{System:} \\
You are an expert-level game player. In the Pendulum game, you control a pendulum attached to a fixed pivot point. The goal is to apply torques to swing the pendulum upright and keep it balanced. The game ends if the pendulum cannot be stabilized within the given time limit. The closer the pendulum is to the upright position, the higher your score. The goal is to swing the pendulum upright and keep it balanced. Provide a torque value (e.g., a float between -2.0 and 2.0) to control the pendulum's movement. Return the torque value enclosed in < and >, e.g., <1.5>.

\vspace{1em}
$\blacktriangleright$ \textbf{User:} \\
The pendulum is at an angle of -2.690 radians from the vertical (zero when upright), rotating at 0.34 radians per second in the clockwise direction.\\
Think step by step.

\vspace{1em}
$\blacktriangleright$ \textbf{Qwen2.5-7B-Instruct:} \\
To determine the appropriate torque to apply, we need to consider the current state of the pendulum: its angle and angular velocity.

\begin{itemize}
    \item \textbf{Angle Analysis}: The pendulum is currently at an angle of -2.690 radians from the vertical. This means it is hanging to the left of the vertical position. To bring it back towards the vertical, we need to apply a positive torque (since positive torque will tend to rotate the pendulum clockwise, which is the direction needed to move it towards the vertical).
    
    \item \textbf{Angular Velocity Analysis}: The pendulum is rotating at 0.34 radians per second in the clockwise direction. Since the pendulum is already moving in the direction we want (towards the vertical), we should continue to apply a positive torque to help it swing more quickly towards the vertical position.
\end{itemize}

Given that the pendulum is already moving in the correct direction but not fast enough, applying a moderate positive torque should help it swing faster and reach the vertical position more efficiently.

A good starting torque value could be around 1.0, as this provides a reasonable amount of force without being too aggressive and risking overshooting the vertical position.

Therefore, the torque value I suggest is: \textbf{<1.0>}
}


\end{document}